\pdfoutput=1

\documentclass[11pt]{article}

\usepackage[preprint]{acl}

\usepackage{times}
\usepackage{latexsym}
\usepackage{hyperref}
\usepackage{url}
\usepackage{multirow} 
\usepackage{graphicx}
\usepackage{colortbl} 

\usepackage{booktabs}
\usepackage{adjustbox}

\usepackage[T1]{fontenc}

\usepackage[utf8]{inputenc}

\usepackage{microtype}

\usepackage{inconsolata}

\usepackage{graphicx}

\usepackage{graphicx} 
\usepackage{makecell}
\usepackage{float}
\usepackage{multirow}
\usepackage{amsmath}
\usepackage{booktabs}
\usepackage{bm}
\usepackage{color}
\usepackage{amsfonts}
\usepackage{subcaption}
\usepackage{marvosym} 

\title{Drawing the Line: Enhancing  Trustworthiness of MLLMs Through the Power of Refusal}%

\author{Yuhao Wang$^1$, 
Zhiyuan Zhu$^1$,
Heyang Liu$^1$, 
Yusheng Liao$^1$, \\
Hongcheng Liu$^1$, 
Yanfeng Wang$^{1,2}$,
Yu~Wang$^{1,2}$\textsuperscript{\Letter}
 \\
  $^{1}$Cooperative Medianet Innovation Center, Shanghai Jiao Tong University \\
  $^{2}$Shanghai Artificial Intelligence Laboratory \\
  \texttt{\{colane, zzysjtu\_iwct, liuheyang, liao20160907, hongcheng\_liu,}\\ \texttt{wangyanfeng622, yuwangsjtu\}@sjtu.edu.cn} \\
}

\begin{document}
\maketitle

\renewcommand{\thefootnote}{}
\footnotemark{}
\footnotetext{\Letter:Corresponding author.}
\renewcommand{\thefootnote}{\arabic{footnote}}

\begin{abstract}

Multimodal large language models (MLLMs) excel at multimodal perception and understanding, yet their tendency to generate hallucinated or inaccurate responses undermines their trustworthiness. Existing methods have largely overlooked the importance of refusal responses as a means of enhancing MLLMs reliability. To bridge this gap, we present the \textbf{In}formation \textbf{Bo}undary-aware \textbf{L}earning Framework (InBoL), a novel approach that empowers MLLMs to refuse to answer user queries when encountering insufficient information. To the best of our knowledge, InBoL is the first framework that systematically defines the conditions under which refusal is appropriate for MLLMs using the concept of information boundaries proposed in our paper. This framework introduces a comprehensive data generation pipeline and tailored training strategies to improve the model’s ability to deliver appropriate refusal responses. To evaluate the trustworthiness of MLLMs, we further propose a user-centric alignment goal along with corresponding metrics. Experimental results demonstrate a significant improvement in refusal accuracy without noticeably compromising the model’s helpfulness, establishing InBoL as a pivotal advancement in building more trustworthy MLLMs.

\end{abstract}

\section{Introduction}

Recent advancements in multimodal large language models (MLLMs) have marked a significant breakthrough in AI research, especially in vision-language tasks~\citep{mckinzie2024mm1, Qwen-VL,tong2024cambrian,fu2024vita,li2024llava, zhang2024internlm}. By integrating visual information with large language models (LLMs), these models have exhibited profound capabilities in multimodal understanding and reasoning, allowing them to perform complex tasks. Despite the impressive progress, MLLMs still face notable challenges. One prominent issue is their tendency to generate factually incorrect or hallucinated content, where models confidently describe non-existent visual elements or provide responses that include incorrect knowledge~\citep{bai2024hallucination,zhong-etal-2024-investigating}. Such hallucinations not only reduce the accuracy of the models but also undermine their truthfulness in practical applications, hindering them from being trustworthy AI assistants.

To improve the trustworthiness of MLLMs, previous works primarily focus on improving multimodal alignment algorithms to enhance the models' perceptual and reasoning capabilities, thereby increasing the truthfulness of their outputs~\citep{yu2024rlhf,yu2024rlaif,amirloo2024understanding}. However, all models have intrinsic limitations in their knowledge and perceptual capabilities, making them prone to produce inaccurate or misleading responses when confronted with tasks beyond their capabilities. Therefore, another effective approach to improving trustworthiness is to train these models to recognize their boundaries and refuse to answer questions when appropriate. While refusal responses may not directly assist the user, they are truthful since no misinformation is provided. 

Despite the critical role of refusal responses, few studies have focused on effectively training MLLMs for this capability. Existing approaches~\citep{liu2023mitigating, cha2024visually} primarily target ambiguous or unanswerable queries, such as those involving non-existent visual elements, but fall short of addressing the broader challenges related to intrinsic limitations and self-awareness in MLLMs~\citep{wang2024mm}. This gap underscores the need for strategies that enable MLLMs to recognize their limitations, ensuring they either provide accurate responses or appropriately refuse to answer when necessary.

{While research on trustworthiness in MLLMs is still limited, efforts to improve reliability by training models to refuse answering unknown questions have been extensively studied in LLMs~\citep{amayuelas-etal-2024-knowledge, yin-etal-2023-large, yang2023alignment, cheng2024can, chen2024teaching, liang2024learning}. These studies typically generate instruction and preference data that include refusal responses, guiding models to avoid answering questions beyond their knowledge boundaries. Trustworthiness is evaluated by examining how well a model can recognize its limitations—providing helpful responses within its knowledge scope and abstaining from answering questions outside of it. However, applying this framework to MLLMs introduces several unique challenges. In multimodal scenarios, trustworthiness depends not only on the model's knowledge but also on its interpretation of visual input and perceptual capabilities, adding complexity to training. Additionally, evaluating trustworthiness requires classifying test questions as `known' or `unknown' based on the model's knowledge boundary—a task complicated by the often ambiguous nature of these boundaries. Moreover, knowledge boundaries vary between models, making consistent comparison of trustworthiness across models using a common, model-agnostic evaluation set particularly challenging.}

{To address these limitations, we propose novel approaches for both training and evaluating the trustworthiness of MLLMs. For model training, we introduce the \textbf{In}formation \textbf{Bo}undary-aware \textbf{L}earning Framework (InBoL), the first to establish the concept of information boundaries and systematically define the conditions under which MLLMs should appropriately refuse to respond. This marks a significant advancement in trustworthiness training for MLLMs. Building on these boundaries, we develop a data construction pipeline that generates `I Don’t Know' (IDK) instruction and preference data from any VQA dataset. Using this data, we implement two key training methods: IDK Instruction Tuning (IDK-IT) and Confidence-aware Direct Preference Optimization (CA-DPO), which enables MLLMs to recognize their information boundaries and refuse to answer when necessary. To evaluate model trustworthiness, we propose a novel alignment objective centered on human preferences rather than intrinsic model metrics. We argue that users consider an MLLM as trustworthy when it provides as many helpful responses as possible while minimizing misinformation. This user-centric approach simplifies evaluation and introduces a model-agnostic framework for assessing trustworthiness. Our experimental results demonstrate that InBoL significantly improves the trustworthiness of baseline models by enhancing their ability to appropriately refuse responses while maintaining helpfulness. This work introduces a new paradigm for developing trustworthy MLLMs and sets the foundation for future advancements in this critical area.}

{Overall, the key contributions of our work are as follows: 
\begin{itemize}
    \item \textbf{InBoL Framework: }We propose the InBoL framework, which introduces the novel concept of information boundaries and integrates a comprehensive data construction pipeline along with tailored training methods. InBoL enhances the trustworthiness of MLLMs by empowering them to recognize these boundaries and refuse to answer when lacking sufficient information, setting a new benchmark for trustworthiness training.

    \item \textbf{User-centric Trustworthiness Evaluation:} We introduce a novel, user-centered alignment objective that shifts the focus of evaluation from model-based metrics to human preferences. This approach simplifies the evaluation process and is model-agnostic. Additionally, we present several metrics that offer a comprehensive and generalizable method for evaluating the trustworthiness of different MLLMs.

    \item \textbf{Experimental Validation: }We conduct extensive experiments to validate the effectiveness of our approach, demonstrating significant improvements in MLLMs' ability to recognize information boundaries while preserving helpfulness. Our detailed analyses offer valuable insights into the broader impact of this method, paving the way for future developments in trustworthy MLLMs.
    
\end{itemize}}

\section{Problem Formulation}

\subsection{MLLM Alignment for Trustworthiness}

Previous studies have explored alignment objectives for trustworthiness in LLMs, primarily focusing on evaluating trustworthiness based on the model’s knowledge boundary. In these works, models are expected to provide accurate and helpful answers when responding to questions within their knowledge scope, and to refuse to answer questions beyond this scope. Formally, given a user query $q$ and the model-generated response $r$, the trustworthiness of this response is evaluated by a value function $v(q, r) \in \{0, 1\}$. The goal of alignment is to maximize $\sum_{q \in D_{\text{test}}} v(q, r)$.

To determine the value of $ v(\cdot) $, these studies first classify the test set questions into known $ D_{k} $ and unknown $ D_{uk}$ categories based on the model's knowledge boundary. The value function $ v(\cdot) $ is then defined as:
\begin{equation}
    v(q,r_{}) = \begin{cases}
1& \text{if } q \in D_{k} \text{ and }r \text{ is correct.} \\
1& \text{if } q \in D_{uk} \text{ and }r \text{ is a refusal response.} \\
0& \text{otherwise}
\end{cases}
\end{equation}

However, categorizing questions as `known' or `unknown' for each model is challenging due to the inherent difficulty of precisely determining a model's knowledge boundary,  which makes the evaluation process complex. Additionally, $D_k$ and $D_{uk}$ are model-specific, making it difficult to fairly compare the trustworthiness of different models.  Moreover, this formulation is also not well-suited for multimodal scenarios. For MLLMs, it is essential to consider not only the model's knowledge but also its perceptual capabilities, given the involvement of visual input.

To address these challenges, we propose a new model-agnostic alignment objective for trustworthiness that is applicable to MLLMs. Inspired by~\citet{xu2024rejection}, our approach evaluates trustworthiness based on user preferences, which can be summarized as follows:

\begin{itemize}
    \item Correct Responses: Users value answers that are accurate, relevant, and informative.
    \item Refusal Responses: Users appreciate refusal responses, as they prevent misinformation and are preferable to incorrect answers.
    \item Incorrect Responses: Users find incorrect answers highly harmful, as they can lead to confusion and misguidance.
\end{itemize}

Based on these preferences, we argue that a trustworthy MLLM should aim to maximize helpful responses while minimizing misinformation. Consequently, the objective of trustworthiness alignment is to train MLLMs to prioritize accuracy and generate refusal responses when necessary to prevent incorrect answers. To reflect this, we redefine the value function as follows:

\begin{equation}
    v(i, q, r) = \begin{cases}
1 & \text{if } r \text{ is a correct response}, \\
0 & \text{if } r \text{ is a refusal response}, \\
-1 & \text{if } r \text{ is a incorrect response}.
\end{cases}
\end{equation}

Consequently, the new objective for trustworthiness alignment is to maximize the sum of values over the test set:

\begin{equation}
\begin{aligned}
    & \underset{\theta}{\text{maximize}} 
    & & \sum_{(i, q) \in D_{\text{test}}} v(i, q, r)
\end{aligned}
\end{equation}
This objective encourages models to generate as many correct responses as possible while prioritizing refusal when accuracy cannot be guaranteed. Unlike previous approaches, the definition of $v(\cdot)$ is model-agnostic, allowing for consistent evaluation across different models, regardless of their intrinsic boundaries. Furthermore, this formulation is more general and can be applied to both unimodal and multimodal scenarios.


\subsection{Evaluation Metrics}

To evaluate the model's trustworthiness, we intoduce two key metrics—Accuracy (Acc) and Refusal Rate (RefR)—defined as follows: 
\begin{equation}
    \text{Acc} = \frac{N_c}{N}, \quad \text{RefR} = \frac{N_r}{N}
\end{equation}
where \( N_c \) is the number of correct responses, \( N_r \) is the number of refusal responses, and \( N \) is the total number of queries.

Combining these two metrics, we define the objective for trustworthiness alignment as  trustworthiness score \( s_{\text{trust}} \) as follows:

\begin{equation}
    \begin{aligned}
        s_{\text{trust}} = \sum_{(i, q) \in D_{\text{test}}} v(i, q, r) = 2 \cdot \text{Acc} + \text{RefR} - 1.
    \end{aligned}
\end{equation}

This score reflects the overall trustworthiness of the model, rewarding both accuracy and refusal responses while penalizing incorrect answers. Unlike previous objectives for LLMs trustworthiness alignment, our evaluation method is simpler and more general.

\begin{figure*}
    \centering
    \includegraphics[width=1\linewidth]{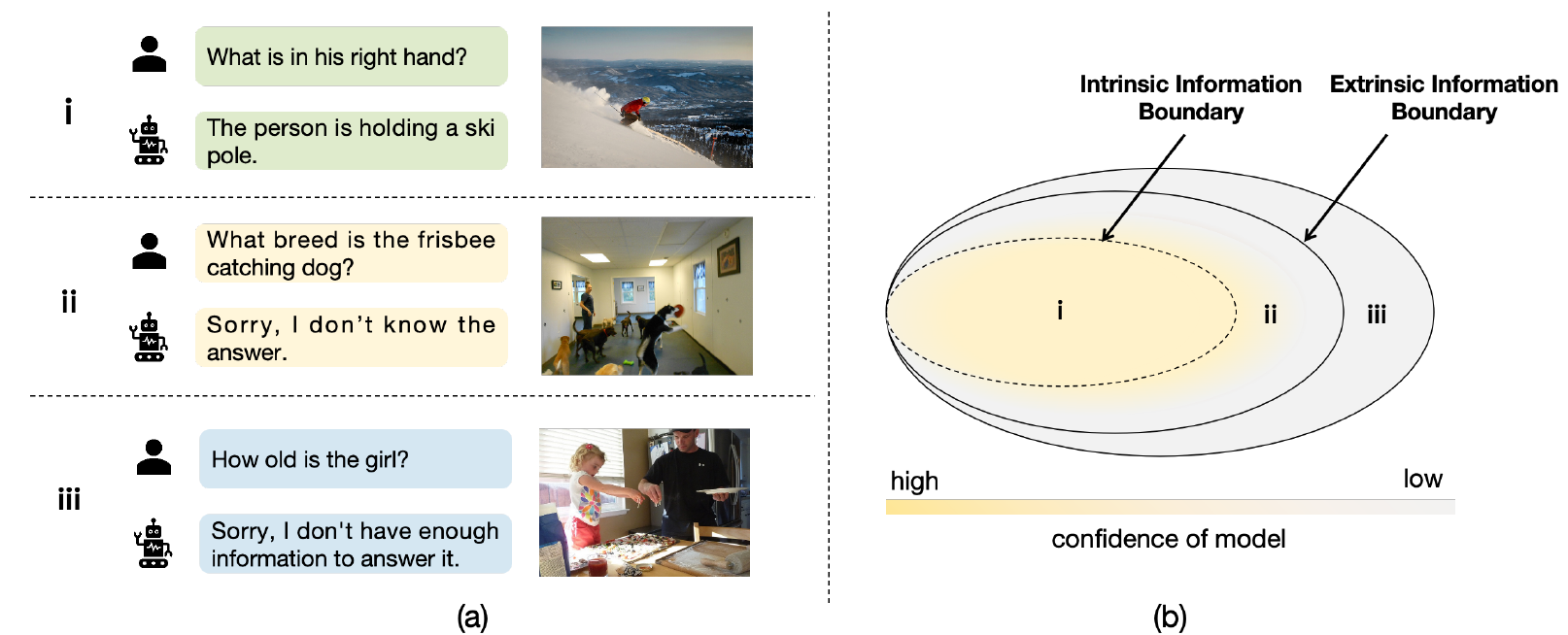}
    \caption{Information Boundaries of MLLMs. (a) Questions are categorized into three types based on intrinsic and extrinsic information boundaries. For Type 1 questions, which fall within the intrinsic boundary, the model is expected to provide helpful responses. For Type 2 questions, which require information unknown to the model, the model should refuse to answer. For Type 3 questions, where the provided image lacks sufficient information, the model should also respond with a refusal. (b) The intrinsic and extrinsic boundaries are illustrated, highlighting the model's varying confidence in answering queries across different regions.}
    \label{fig:info-bound}
\end{figure*}

\section{Information Boundary-Aware Learning framework}
To enhance the trustworthiness of MLLMs, we propose the \textbf{In}formation \textbf{Bo}undary-Aware \textbf{L}earning Framework (InBoL). This framework includes a data construction pipeline designed to generate model-specific `IDK' instruction and preference data by considering the intrinsic and extrinsic information boundaries of MLLMs. Furthermore, we incorporate `IDK' instruction tuning (IDK-IT) and confidence-aware direct preference optimization (CA-DPO) for model training. The goal of this framework is to improve the model's ability to provide appropriate refusal responses, thereby reducing misinformation and increasing reliability of MLLMs.

\subsection{Information Boundary}
\label{sec:info_bound}
The core of our framework is to train MLLMs to recognize when to refuse, thereby avoiding the generation of misinformation. While previous work on LLMs generally restricts refusals to questions outside the model's knowledge boundary, multimodal scenarios introduce additional complexity, as both visual information and knowledge must be considered.

To address this, we introduce extrinsic and intrinsic information boundaries for MLLMs, as illustrated in Figure~\ref{fig:info-bound}. In multimodal scenarios, a trustworthy MLLM should answer questions only when it has sufficient information and refuse when it does not, and these  boundaries serve as guidelines for this decision-making process. 

\paragraph{Extrinsic Information Boundary} In multimodal scenarios, MLLMs depend on extrinsic visual inputs to respond to user queries. The extrinsic information boundary defines the distinction between what is explicitly present in the visual input and what is absent. When the necessary information to answer a question is not available—indicating that the query exceeds the extrinsic boundary—the model should provide a refusal response.

\paragraph{Intrinsic infomration boundary} Beyond the extrinsic boundary, a model's intrinsic information boundary is equally important, defined by its inherent capabilities. This boundary encompasses what the model can infer from the image and the multimodal knowledge embedded in its parameters. If the model cannot perceive the required information from the image or lacks specific knowledge, thereby exceeding the intrinsic boundary, it should also provide a refusal response.


\begin{figure*}
    \centering
    \includegraphics[width=1\linewidth]{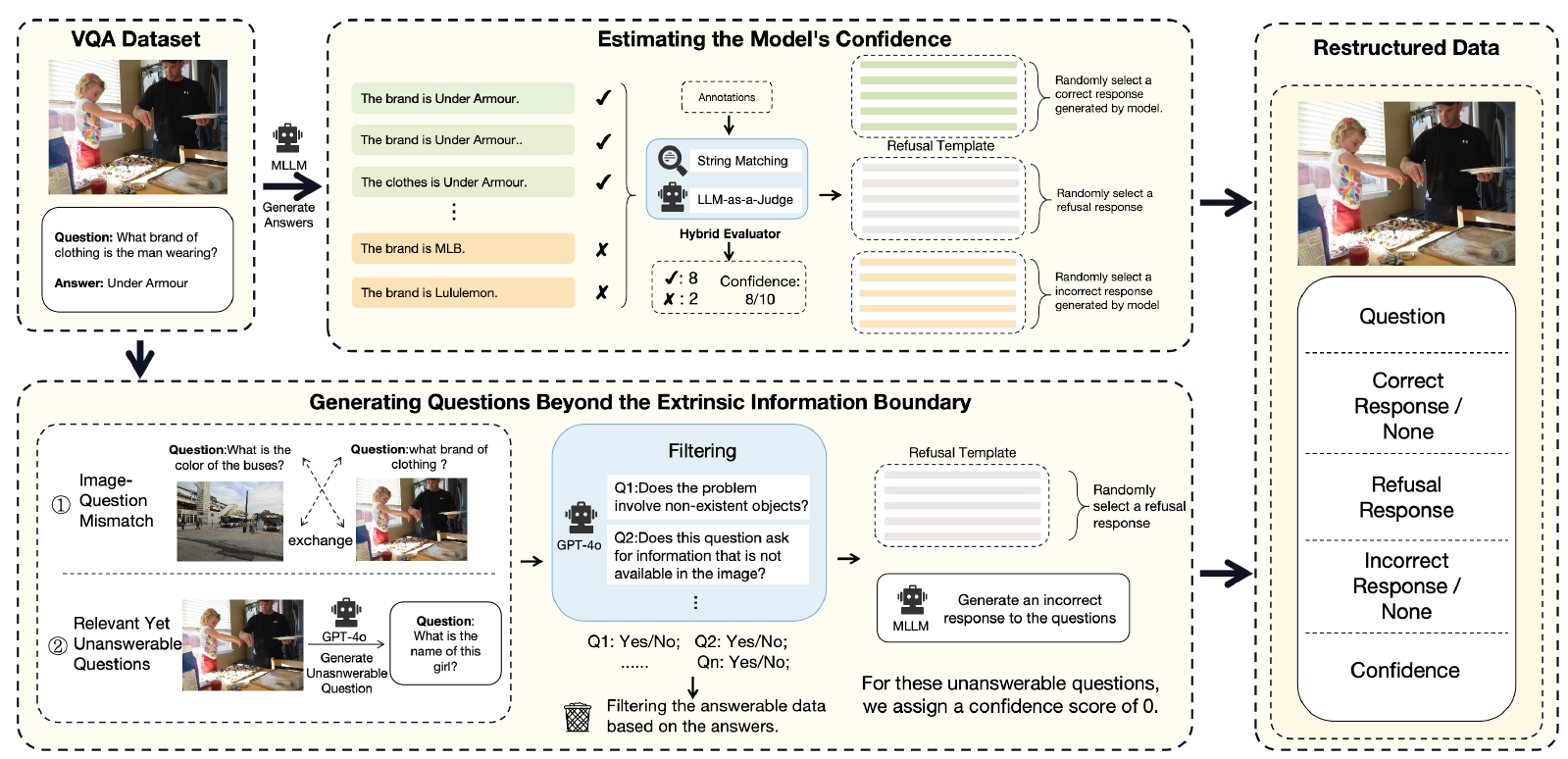}
    \caption{The Pipeline of Data Construction: Given a VQA dataset, we design a pipeline to collect different types of samples within and beyond the information boundaries. First, we estimate the confidence for each sample to determine the model's intrinsic information boundary. Next, we generate questions that lie beyond the extrinsic boundary, followed by quality filtering. Finally, all data is formatted into a standardized structure, including correct, incorrect, and refusal responses, each accompanied by their corresponding confidence scores.}
    \label{fig:data_construction}
\end{figure*}

\subsection{Data Construction}
\label{sec:data_cons}


To train MLLMs to appropriately refuse questions, we need to collect the three types of VQA data outlined in Figure~\ref{fig:info-bound}. For any given VQA dataset, we propose a data construction pipeline that classifies questions into the first two types based on confidence estimation, while generating unanswerable questions as the third type from the available data. Additionally, we reorganize the generated data into a standardized format, as shown in Figure~\ref{fig:data_construction}, which is then used to create the `IDK' instructions and preference data.

\vspace{-0.2cm}
\paragraph{Estimating the Model's Confidence}
We assume that all questions in a given VQA dataset are answerable based on the provided visual information, placing them within the extrinsic information boundary. To further determine if these questions fall within the intrinsic information boundary, we estimate the model's confidence. Following prior works~\citep{cheng2024can, xu2024rejection, yang2023alignment}, we sample multiple responses from the original model and calculate the accuracy rate to estimate its confidence. We found that simple string matching—checking if the correct answers appear in the generated responses—was insufficient, as the model sometimes generates semantically similar answers that differ in wording. To address this issue, we develop a hybrid evaluator by utilizing a LLM to evaluate responses marked as incorrect by the string-matching method. More details about it can be found in Appendix~\ref{sec:Hybrid_Evaluator}. In addition, we randomly sample a correct and incorrect answers from the model-generated responses and select a refusal response from the refusal template  (Appendix~\ref{sec:Refusal_Template}), then restructure the data into a standardized format as illustrated in  Figure~\ref{fig:data_construction}.


\vspace{-0.2cm}
\paragraph{Generating Unanswerable Questions}
Next, we generate questions that lie beyond the extrinsic information boundary. First, we focus on questions that are irrelevant to the provided image, such as those inquiring about non-existent objects. To create these, we randomly select samples from the VQA dataset and reorganize them into mismatched image-question pairs. Additionally, we formulate more complex questions that, while related to the image, cannot be answered due to incorrect assumptions or insufficient information provided by the image. To achieve this, we design prompts that instruct GPT-4o to generate unanswerable queries. To ensure the quality of these generated questions, we implement a filtering mechanism. Specifically, we define several criteria for determining whether a question is unanswerable and instruct GPT-4o to verify if the questions meet these criteria. If the model responds ``no" to all relevant checks, the question is excluded from the dataset. For each generated unanswerable question, we assign a confidence score of 0. Additionally, we collect an incorrect response generated by the original model and select a refusal response from predefined templates to restructure the data into the standardized format shown in Figure~\ref{fig:data_construction}. Detailed descriptions of the generation and filtering processes are provided in Appendix~\ref{sec:ua_gen} and we further discuss the generalization of this data generation method in Appendix~\ref{sec:gdg}.

\begin{figure*}

    \centering
    \includegraphics[width=1\linewidth]{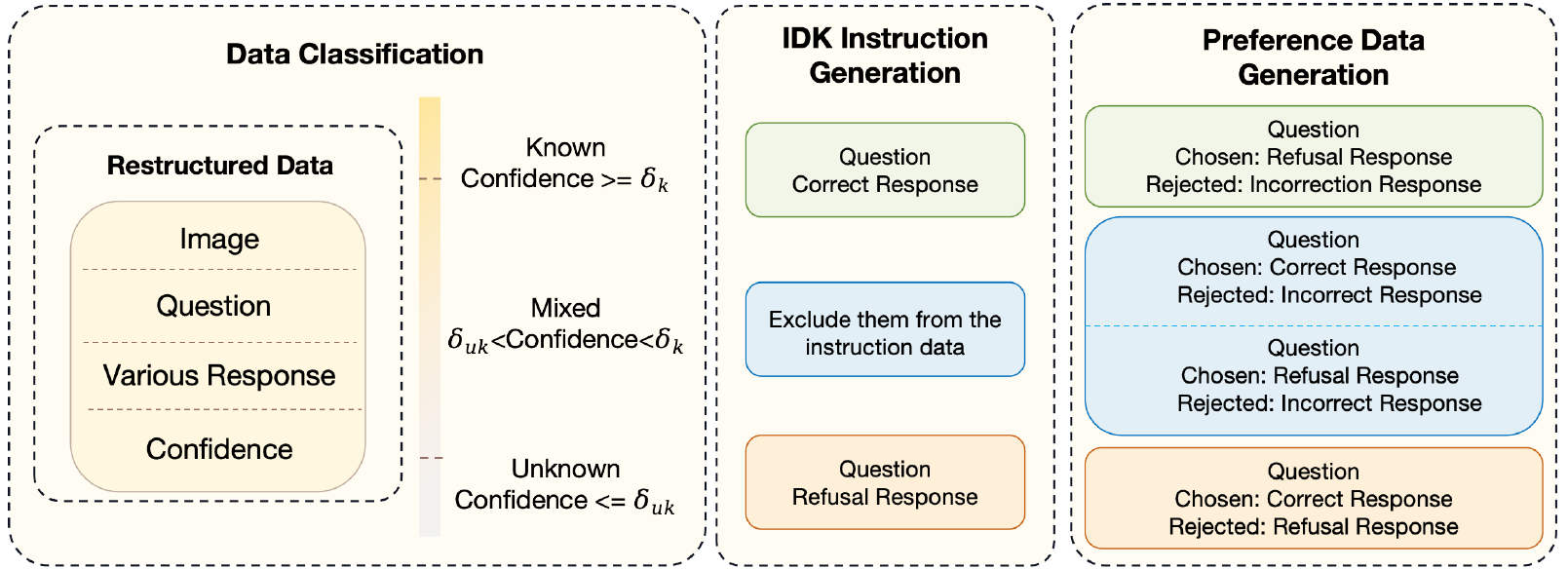}
    \caption{Construction of `IDK' instruction and preference data: The restructured data is categorized into `Known,' `Mixed,' and `Unknown' based on confidence thresholds($\delta_{k}$ and $\delta_{uk}$). `IDK' instruction generation includes correct responses for known questions, refusal responses for unknown questions, and the exclusion of mixed data. Preference data samples are constructed by pairing questions with correct, incorrect, and refusal responses, based on the confidence classification of each question.}
    \label{fig:instruction}
\end{figure*}

\vspace{-0.2cm}
\paragraph{Constructing `IDK' Instruction}
To construct the `IDK' instruction, we categorize the restructured data into three types  based on the confidence thresholds $\delta_k$ and $\delta_{uk}$: `Known,' `Mixed,' and `Unknown,' as shown in Figure~\ref{fig:instruction}. For known questions, we select the correct answer as the response. For unknown questions, we utilize the refusal response. Regarding the `Mixed' data, we exclude it from the instruction data, as the model exhibits relatively high uncertainty for these questions.

\vspace{-0.2cm}
\paragraph{Constructing Preference Data}
A preference data consists of a question, a chosen response, and a rejected response. For known questions, we use the correct answers as the chosen response and the refusal as the rejected response. For unknown questions, we utilize the refusal answers as the chosen response and the incorrect answers as the rejected response. For mixed questions, we construct two samples, both of which use incorrect answers as the rejected response. In one sample, the chosen response is the correct answer, while in the other, the chosen response is the refusal.


\subsection{Model Training for Information Boundary Awareness}

To enhance the model's ability to recognize and refuse questions beyond its information boundary, we propose two training strategies: `IDK' Instruction Tuning (IDK-IT) and Confidence-aware Direct Preference Optimization (CA-DPO). These strategies can teach the model not only to provide accurate responses but also to give refusal responses when it lacks the necessary information.

\vspace{-0.2cm}
\paragraph{IDK Instruction Tuning}
Instruction tuning is an effective method for aligning the model's responses with desired behaviors.  In our framework, we train the model with `IDK' instructions. This training approach improves models trustworthiness by reducing the generating misinformation.

\vspace{-0.2cm}
\paragraph{Confidence-aware Direct Preference Optimization}
Direct Preference Optimization  (DPO) is a technique that optimizes a model's policy using preference data~\citep{xu2023some,rafailov2024direct, hong2024reference,yuan2024self}. While DPO can guide models to prefer correct answers and learn to refuse when needed, it does not leverage the model's intrinsic confidence to dynamically adjust its behavior. To address this, we propose Confidence-aware DPO (CA-DPO), which integrates the model's confidence score into the optimization process.

As shown in Figure~\ref{fig:instruction}, we define two preference pairs for `Mixed' samples: $p_1$ (correct $>$ incorrect) and $p_2$ (refusal $>$ incorrect). For consistency, we define `Known' samples with $p_1 = p_2 =$ (correct $>$ refusal), and `Unknown' samples with $p_1 = p_2 =$ (refusal $>$ incorrect). Our approach uses the confidence score to dynamically balance the emphasis between these preference pairs. The CA-DPO loss function is defined as: 


\begin{equation}
\begin{aligned}
    f(x, p) =  \log \sigma ( \beta \log \frac{\pi_*(y_{w} | x)}{\pi_{\text{ref}}(y_{w} | x)} - \beta \log \frac{\pi_*(y_{l} | x)}{\pi_{\text{ref}}(y_{l} | x)} )
\end{aligned}
\end{equation}

\begin{equation}
    \begin{aligned}
    \mathcal{L}_{\text{cadpo}} = & -\mathbb{E}_{(x,p_1,p_2)} \Big(
    f(x, p_1) \cdot {conf}_x +\\
    & f(x, p_2) \cdot (1 - {conf}_x)
    \Big)
    \end{aligned}
\end{equation}

where $(y_{w1} > y_{l1})$ represents $p_1$, $(y_{w2} > y_{l2})$ represents $p_2$, and $conf_{x}$ is the model's confidence score. The confidence score adjusts the balance between the two preference pairs, particularly for `Mixed' samples. In high-confidence scenarios, the loss function prioritizes correct responses, while in low-confidence cases, it favors refusal. This adaptive mechanism enables the model to balance cautiousness and helpfulness more effectively.


\section{Experimental Setup}

\subsection{Training Data}

As mentioned in Section~\ref{sec:info_bound}, our work considers both the model's knowledge and visual information. Therefore, we use general VQA datasets and knowledge-intensive VQA datasets for data construction. Specifically, we utilize VQAV2~\citep{antol2015vqa, zhang2016yin, goyal2017making}, Oven~\citep{hu2023open}, and ScienceQA~\citep{lu2022learn}. We set the confidence thresholds as $\delta_{k}=0.8$ and $\delta_{uk}=0.2$. For `IDK' instruction tuning, we collected  11k instructions. For CA-DPO, we gathered about 24k preference pairs. Further training details are provided in Appendix~\ref{sec:train_detail}.



\subsection{Evaluation}

In our experiments, we
utilize the LLaVA1.5~\citep{liu2023llava,liu2023improvedllava}  model, one of the most widely used open-source MLLMs. We evaluate models on both in-domain and out-of-domain (OOD) datasets. For the in-domain evaluation, we draw questions from the validation sets of VQAV2 and Oven, as well as the test set of ScienceQA. In addition, we generate unanswerable questions (UaVQA) as described in Section~\ref{sec:data_cons} and manually filter them for evaluation purposes. The final in-domain dataset consists of 1,000 samples.

For the OOD evaluation, we assess the model on three types of benchmarks: general VQA, knowledge-intensive VQA, and unanswerable VQA. For general VQA, we use the AOKVQA validation set~\citep{schwenk2022okvqa}, the GQA test set~\citep{hudson2019gqa}, and the MMBench (en-dev)~\citep{liu2023mmbench}. In the case of knowledge-intensive VQA, we employ the validation set of MMMU~\citep{yue2023mmmu}. For unanswerable VQA, we adopt the BeyondVisQA subset from MM-SAP~\citep{wang2024mm}. 

\subsection{Baselines}
In terms of baselines, we consider both prompt-based and training-based methods. Refusal Prompt instruct the model to refuse answering when it lacks sufficient information by appending a prompt to the text input. The refusal prompt is: If you don't have enough information to answer the question, respond with ``Sorry, I can not help with it.'' We also conduct supervised fine-tuning(SFT) as baseline. For questions within the extrinsic information boundary, the model is trained using the correct answers. For questions outside this boundary, since no correct answers exist, we assign an `IDK' response as the label. By constructing the dataset in this manner, we fine-tune models with about 11k instructions. Further details about evaluation can be found in the Appendix~\ref{sec:evaluation}.

\vspace{-0.8mm}

\begin{table*}[t]
\begin{center}
\resizebox{1.0\textwidth}{!}{%
\begin{tabular}{lcccp{1pt}cccp{1pt}ccccccc}
\hline

\multirow{2}{*}{\textbf{Method}}
 & \multicolumn{3}{c}{\textbf{AOKVQA}} & & \multicolumn{3}{c}{\textbf{GQA}} & & \multicolumn{3}{c}{\textbf{MMMU}} & \textbf{BeyondVisQA} & \multicolumn{3}{c}{\textbf{MMBench(en-dev)}} \\ 
\arrayrulecolor[gray]{0.5}
\cline{2-4}  \cline{6-8}  \cline{10-12} \cline{14-16 }
\arrayrulecolor{black}

 & {Acc} & {RefR} & {$S_{trust}$} & & {Acc} & {RefR} & {$S_{trust}$} & & {Acc} & {RefR} & {$S_{trust}$} & {RefR} & {Acc} & {RefR} & {$S_{trust}$} \\
\hline
\textbf{LLaVA1.5-7B}  & 78.56  & 0.00  & 57.13 & & 59.65  & 0.00  & 19.30 & &  34.70  & 0.00  & -30.60  & 25.50  & 62.80  & 0.00  & 25.60  \\
{+Refusal Prompt} & 56.77  & 26.20  & 39.74 & & 58.65  & 3.43  & 20.74 & & 32.22  & 12.89  & -22.67  & 27.50  & 59.36  & 0.69  & 19.42 \\
{+SFT} & 74.32  & 3.49  & 52.14 & & 59.39  & 2.77  & 21.55 & &  34.20  & 1.67  & -29.93  & 56.00  & 63.32  & 0.26  & 26.89  \\
{+IDK-IT} & 55.50  & 36.24  & 47.24 & & 50.46  & 23.88  & 24.81 & &  15.22  & 69.67  & \textbf{0.11}  & \textbf{75.25}  & 46.39  & 39.09  & 31.87 \\
{+CA-DPO} & 72.23  & 17.64  & \textbf{62.10} & & 60.41  & 12.95  & \textbf{33.77} & & 19.67  & 56.67  & -4.00  & 67.75  & 58.42  & 18.13  & \textbf{34.97} \\
\hline
\textbf{LLaVA1.5-13B} & 78.95  & 0.00  & 57.90 & & 61.81  & 0.00  & 23.63 & &  36.22  & 0.00  & -27.56  & 33.50  & 67.96  & 0.00  & 35.91 \\
{+Refusal Prompt} & 63.32  & 18.95  & 45.59 & & 61.36  & 1.96  & 24.69 & & 27.78  & 19.56  & -24.89  & 46.00  & 64.69  & 0.26  & 29.64 \\
{+SFT} & 77.82  & 2.62  & 58.25 & & 61.32  & 1.69  & 24.33 & & 38.22  & 1.78  & -21.78  & 68.75  & 67.01  & 0.00  & 34.02  \\
{+IDK-IT} & 63.93  & 23.06  & 50.92 & & 52.27  & 19.22  & 23.77 & &  14.22  & 74.33  & \textbf{2.78}  & \textbf{79.50}  & 55.84  & 23.91  & 35.60   \\
{+CA-DPO} & 73.89  & 15.63  & \textbf{63.41} & & 59.70  & 13.82  & \textbf{33.22} & & 25.89  & 41.78  & -6.44  & 72.50  & 62.63  & 14.69  & \textbf{39.95}  \\
\hline
\end{tabular}%
}
\end{center}
\caption{Performance on out-of-domain dataset. We present results on LLaVA1.5-7B and LLaVA1.5-13B. Bold values indicate the highest trustworthiness score.}
\label{tab:ood}
\end{table*}

\section{Results and Analysis}

\begin{table}[h]
\begin{center}
\resizebox{0.4\textwidth}{!}{%
\begin{tabular}{lccc}
\hline
\textbf{Method} 
 & {Acc} & {RefR} & {$S_{trust}$} \\
\hline
\textbf{LLaVA1.5-7B}  & 12.00  & 46.10  & -6.50 \\
{+Refusal Prompt} & 47.00  & 41.70  & -4.10 \\
{+SFT} & 81.00  & 49.10  & 8.90 \\
{+IDK-IT} & \textbf{92.00}  & 38.60  & 16.00 \\
{+CA-DPO} & 87.00  & 49.10  & \textbf{28.50} \\
\hline
\textbf{LLaVA1.5-13B} & 20.00  & 51.00  & 4.10 \\
{+Refusal Prompt}  & 62.00  & 48.20  & 10.80 \\
{+SFT} & 78.00  & 53.60  & 17.30  \\
{+IDK-IT} & 89.00  & 41.80  & 21.20   \\
{+CA-DPO} & \textbf{93.00}  & 49.00  & \textbf{32.00} \\
\hline
\end{tabular}%
}
\end{center}
\caption{Performance on in-domain dataset. We present results on LLaVA1.5-7B and LLaVA1.5-13B. Bold values indicate the highest trustworthiness score.}
\label{tab:in-domain}
\end{table}

\subsection{Overall Results}
\label{main-results}

The results on the in-domain datasets are presented in Table~\ref{tab:in-domain}. Both IDK-IT and CA-DPO demonstrate notable improvements in trustworthiness scores compared to the baselines. IDK-IT increases the model’s refusal rate, enabling it to recognize when sufficient information is lacking. Although IDK-IT does result in a decline in accuracy, it improves the model's trustworthiness by prioritizing cautiousness over potential overconfidence. In contrast, CA-DPO achieves a more balanced outcome by improving the refusal rate while maintaining model accuracy. This suggests that CA-DPO enables the model to better distinguish between known and unknown queries without sacrificing helpfulness. As illustrated in Table~\ref{tab:ood}, IDK-IT and CA-DPO generalize well to OOD datasets. IDK-IT continues to significantly boost the refusal rate to improve the trustworthiness score, while CA-DPO strikes an effective balance between improving the refusal rate and preserving accuracy.

In summary, both IDK-IT and CA-DPO clearly enhance the trustworthiness of models. IDK-IT is particularly effective at reducing misinformation by increasing the refusal rate, though it may make models overly cautious. CA-DPO, meanwhile, achieves a more favorable balance between truthfulness and helpfulness, making models more trustworthy.

\subsection{Awareness of the Extrinsic Information Boundary}
To comprehensively evaluate the model's awareness of the extrinsic information boundary, we conduct additional experiments on the unanswerable subset of VizWiz~\citep{gurari2018vizwiz} and the validation set of VQAv2-IDK~\citep{cha2024visually}. VQAv2-IDK comprises questions from VQAv2 annotated with `IDK' keywords. However, we observe that some of these questions, while challenging, can still be answered based on image information, suggesting that they remain within the extrinsic information boundary. Consequently, we filter out questions that contain only a single `IDK' annotation. VizWiz is a VQA dataset that includes visual questions posed by people who are blind. We select data labeled as `unanswerable' to form its unanswerable subset. As shown in Table~\ref{tab:eib}, our models appropriately provides refusal responses on these out-of-domain datasets, demonstrating their clear awareness of the extrinsic information boundary.

\begin{table*}[t]
\label{sample-table}
\begin{center}
\resizebox{1.0\textwidth}{!}{%
\begin{tabular}{lcccp{1pt}ccc}
\hline
\multirow{2}{*}{}
 & \multicolumn{3}{c}{{LLaVA1.5-7b}} & & \multicolumn{3}{c}{{LLaVA1.5-13b}} \\ 
\arrayrulecolor[gray]{0.5}
\cline{2-4}  \cline{6-8} 
\arrayrulecolor{black} 
 & {Original} & {IDK-IT} & {CA-DPO} &  & {Original} & {IDK-IT} & {CA-DPO} \\
\hline
{Vizwiz(ua)}  & 9.00 & 76.01 & 69.97 & & 9.60 & \textbf{78.61} & 73.27  \\
{VQAv2-IDK(filter)}  & 2.80 & \textbf{81.42} & 70.63 & & 2.60 & 80.14 & 72.40  \\
\hline
\end{tabular}%
}
\end{center}
\caption{Refusal Rate on unanswerable VQA datasets. VizWiz(ua) refers to the unanswerable subset of the VizWiz dataset, while VQAv2-IDK(filter) represents the filtered subset of VQAv2-IDK, where only questions with more than one 'IDK' annotation are retained.}
\label{tab:eib}
\end{table*}

\begin{figure*}
    \centering
    \includegraphics[width=1.0\linewidth]{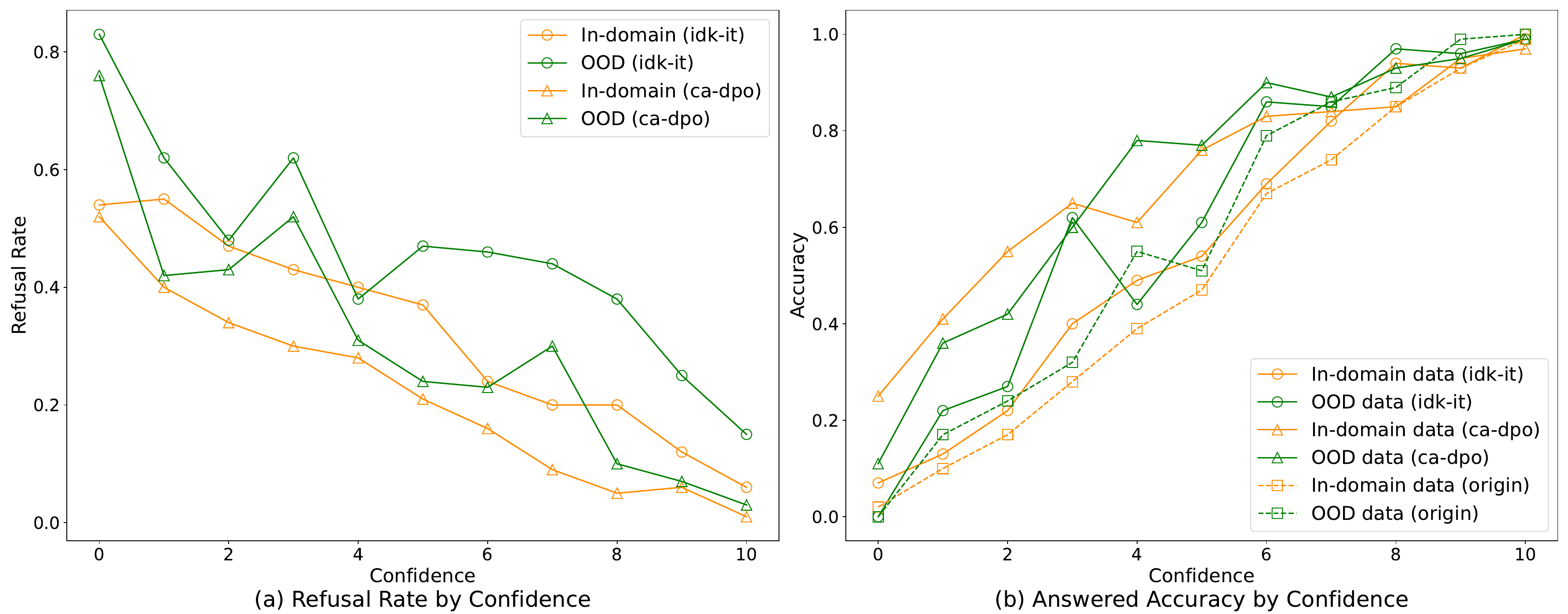}
    \caption{Refusal rate and accuracy of models across different confidence levels. (a) Refusal Rate by Confidence: The model exhibits dynamic refusal behavior, with higher refusal rates for lower confidence levels and a tendency to answer directly for high-confidence questions. This indicates the model's awareness of its intrinsic information boundary. (b) Answered Accuracy by Confidence: The accuracy of the IDK-IT and CA-DPO models surpasses that of the original model, demonstrating that training methods focused on intrinsic boundary recognition improve the model's ability to provide accurate responses when choosing to answer. }
    \label{fig:iib-aware}
\end{figure*}


\begin{table*}[t]
\begin{center}
\resizebox{1.0\textwidth}{!}{%
\begin{tabular}{llccccp{1pt}ccc}
\hline
\multirow{2}{*}{Model} & \multirow{2}{*}{Method}
 & \multirow{2}{*}{Data}
 & \multicolumn{3}{c}{{In-Domain(Avg)}} & & \multicolumn{3}{c}{{Out-Of-Domain(Avg)}} \\ 
\arrayrulecolor[gray]{0.5}
\cline{4-6}  \cline{8-10} 
\arrayrulecolor{black} &
& & {Acc} & {RefR} & {$S_{trust}$} &  & {Acc} & {RefR} & {$S_{trust}$} \\
\hline
\multirow{4}{*}{LLaVA1.5-7B} & {DPO} & (1) & 47.50	&30.20	&25.20 & & 50.30	&29.46 &	30.06  \\
&{DPO}& (2)  & 51.00	&26.30	&28.30 & & 55.08	&18.62 &	28.78  \\
&{DPO}& (3)  & 49.50	&26.30	&25.30 & & 52.88	&20.35&	26.11  \\
&{CA-DPO}& (3)  & 49.10	&30.30	&\textbf{28.50} & & 52.68	&26.35 &	\textbf{31.71}  \\
\hline
\multirow{4}{*}{LLaVA1.5-13B} & {DPO} & (1) & 47.10	&36.10	&30.30 & & 52.71	&24.14 &	29.56 \\
&{DPO}& (2)  & 49.60	&31.40	&30.60 & & 56.37	&16.45 &	29.20  \\
&{DPO}& (3)  & 48.60	&33.90	&31.10 & & 55.19	&20.83&	31.21  \\
&{CA-DPO}& (3)  & 49.00	&34.00	&\textbf{32.00} & & 55.53	&21.48 &	\textbf{32.53}  \\
\hline
\end{tabular}%
}
\end{center}
\caption{Performance comparison between models trained with different preference data using DPO and CA-DPO methods.}
\label{tab:ca-dpo}
\end{table*}

\subsection{Awareness of the Intrinsic Information Boundary}


Although our overall results demonstrate that the proposed training method significantly reduces misinformation by promoting refusals while maintaining strong performance, we aim to further investigate the model's intrinsic awareness of its information boundaries. To this end, we analyze changes in the refusal rates of both IDK-IT and CA-DPO models relative to the confidence levels of LLaVA1.5-7B on both an in-domain dataset and the AOKVQA (OOD) dataset, as illustrated in Figure~\ref{fig:iib-aware}(a). The results indicate that the model exhibits an awareness of its own confidence, effectively refusing to answer when appropriate. For high-confidence questions, the model typically provides direct answers, while for lower-confidence questions, it demonstrates a higher likelihood of refusal. This adaptive refusal behavior reflects the model's capacity to distinguish between instances where it possesses sufficient knowledge and those where it does not, underscoring its intrinsic self-awareness.



Additionally, we calculate the accuracy of the answered questions, defined as: $\text{Answered Acc} = \frac{N_c}{N-N_r}$, where $N_c$ is the number of correct answers and $N_r$ is the number of refusals. Figure~\ref{fig:iib-aware}(b) shows the answered accuracy for the original model, the IDK-IT model, and the CA-DPO model. Notably, the accuracy curve for the original model is lower than those of both the IDK-IT and CA-DPO models. This indicates that training the model to recognize its intrinsic information boundaries through the IDK-IT and CA-DPO methods enhances its ability to more effectively utilize the information it possesses, leading to improved overall accuracy.

Thus, training the model to be aware of its information boundaries improves trustworthiness in two key ways. First, by identifying these boundaries, the model learns to refuse answering when it lacks sufficient information, significantly reducing the risk of generating misinformation. Second, this awareness allows the model to provide more accurate responses when it does choose to answer, as it makes more effective use of its knowledge. Together, these improvements contribute to a more reliable and trustworthy MLLM.

\subsection{The effectiveness of CA-DPO}
To evaluate the effectiveness of CA-DPO, we used three types of preference data for the `mixed' samples and trained LLaVA1.5 using the original DPO loss. These preference pairs include: (1) refusal $>$ incorrect; (2) correct $>$ incorrect; and (3) a combination of both, which also serves as the training data for CA-DPO. Table~\ref{tab:ca-dpo} presents the average performance on both in-domain and OOD datasets. The results indicate that models trained with the CA-DPO loss achieve a more balanced performance between accuracy and refusal rate, resulting in the highest trustworthiness score. This suggests that the CA-DPO method encourages the model to be more selective in its responses, striking an effective balance between providing helpful answers and refusing when necessary, thereby enhancing its overall trustworthiness.


\subsection{Distinguishing Intrinsic and Extrinsic Information Deficits}
As mentioned in Section~\ref{sec:info_bound}, we categorize unknown questions into two types based on whether the model fails to answer due to a lack of visual information or internal knowledge. A key question is whether the model can differentiate between these two cases—that is, whether it can identify the specific type of information it lacks. To investigate this question, we apply linear probing to explore whether the model’s internal representations encode features that differentiate between these two types of unknown cases. 

Linear probing has been widely used in understanding and extracting knowledge of LLMs and MLLMs~\citep{zhao2025first, qian2024towards}. We selected 2,000 unknown questions from the training dataset with confidence score below 2, which are beyond extrinsic or intrinsic boundaries. For each question, we fed it into the MLLM and extracted the first generated token from the final layer of the model. We then trained a three-layer linear classifier to categorize these tokens into two classes. To evaluate the classifier,  we used a set of 200 unknown questions (confidence scores below 2) from the in-domain test set. The classification accuracy of the linear probing is shown in Table~\ref{tab:classification_accuracy}.

The results demonstrate that a simple linear classifier achieves high accuracy in distinguishing between intrinsic and extrinsic deficits based on the first generated token of MLLMs. This suggests that the model’s internal representations encode features that effectively differentiate between these two types of unknown questions. This finding indicates that the model has the potential to recognize the source of missing information, even though its explicit refusal responses do not currently articulate this distinction. Incorporating this capability into the model’s outputs could further enhance trustworthiness by providing more transparent refusals. Future work could leverage this potential by introducing mechanisms for explanation behind refusals, thereby aligning the model’s behavior more closely with user expectations for trustworthy AI systems.
\begin{table}[t]
\label{sample-table}
\begin{center}
\resizebox{0.5\textwidth}{!}{%
\begin{tabular}{lccp{1pt}cc}
\hline
\multirow{2}{*}{}
 & \multicolumn{2}{c}{{GPT-4o-generated data}} & & \multicolumn{2}{c}{{Qwen2-VL-generated data}} \\ 
\arrayrulecolor[gray]{0.5}
\cline{2-3}  \cline{5-6} 
\arrayrulecolor{black} 
  & {IDK-IT} & {CA-DPO} &   & {IDK-IT} & {CA-DPO} \\
\hline
{Vizwiz(ua)}   & \textbf{76.01} & 69.97 &  & {74.49} &  71.39 \\
{VQAv2-IDK(filter)}  & \textbf{81.42} & 70.63 & & 79.25 &  75.22 \\
{BeyondVisQA}  & \textbf{75.25} & 67.75 & & 72.50 & 69.50  \\
\hline
\end{tabular}%
}
\end{center}
\caption{Performance on unanswerable VQA datasets using GPT-4o and Qwen2-VL-72B for data generation.Bold values highlight the highest refusal rate for each dataset.}
\label{tab:qwenua}
\end{table}

\section{Conclusion}
In this paper, we introduce the InBoL Framework to enhance the trustworthiness of MLLMs. By defining information boundaries, we create a data generation pipeline and apply novel training methods—IDK-IT and CA-DPO—to improve models' ability to avoid misinformation while maintaining helpfulness. Our user-centric evaluation approach also offers a model-agnostic way to assess trustworthiness. Experimental results show that our method effectively reduces misinformation and enhances model reliability, paving a feasible path for the future development of trustworthy MLLMs.

\section*{Limitations}
In this work, we did not explore the generation of explanations for refusal responses, an important and underexamined area. From the model's perspective, many questions require reasoning processes to determine whether sufficient information is available to provide an accurate answer. By incorporating explanations for refusal responses, the model could better learn when to refuse appropriately, thereby enhancing its awareness of its own limitations and boundaries. From the user's perspective, unexplained refusals may lead to confusion or dissatisfaction. Providing clear and interpretable justifications for refusals could make the refusal mechanism more transparent and user-friendly, significantly improving the overall user experience.

For future work, we plan to focus on enabling the model to generate well-reasoned and contextually appropriate refusal explanations. This will involve developing methodologies for constructing relevant datasets and designing robust evaluation frameworks to assess the quality and relevance of the generated explanations. By making refusal responses more informative and transparent, we aim to further enhance the trustworthiness of the model while ensuring a more positive and engaging user experience.

\bibliography{custom}

\begin{thebibliography}{46}
\providecommand{\natexlab}[1]{#1}

\bibitem[{Amayuelas et~al.(2024)Amayuelas, Wong, Pan, Chen, and Wang}]{amayuelas-etal-2024-knowledge}
Alfonso Amayuelas, Kyle Wong, Liangming Pan, Wenhu Chen, and William~Yang Wang. 2024.
\newblock \href {https://aclanthology.org/2024.findings-acl.383} {Knowledge of knowledge: Exploring known-unknowns uncertainty with large language models}.
\newblock In \emph{Findings of the Association for Computational Linguistics ACL 2024}, pages 6416--6432, Bangkok, Thailand and virtual meeting. Association for Computational Linguistics.

\bibitem[{Amirloo et~al.(2024)Amirloo, Fauconnier, Roesmann, Kerl, Boney, Qian, Wang, Dehghan, Yang, Gan et~al.}]{amirloo2024understanding}
Elmira Amirloo, Jean-Philippe Fauconnier, Christoph Roesmann, Christian Kerl, Rinu Boney, Yusu Qian, Zirui Wang, Afshin Dehghan, Yinfei Yang, Zhe Gan, et~al. 2024.
\newblock Understanding alignment in multimodal llms: A comprehensive study.
\newblock \emph{arXiv preprint arXiv:2407.02477}.

\bibitem[{Antol et~al.(2015)Antol, Agrawal, Lu, Mitchell, Batra, Zitnick, and Parikh}]{antol2015vqa}
Stanislaw Antol, Aishwarya Agrawal, Jiasen Lu, Margaret Mitchell, Dhruv Batra, C~Lawrence Zitnick, and Devi Parikh. 2015.
\newblock Vqa: Visual question answering.
\newblock In \emph{Proceedings of the IEEE international conference on computer vision}, pages 2425--2433.

\bibitem[{Bai et~al.(2023)Bai, Bai, Yang, Wang, Tan, Wang, Lin, Zhou, and Zhou}]{Qwen-VL}
Jinze Bai, Shuai Bai, Shusheng Yang, Shijie Wang, Sinan Tan, Peng Wang, Junyang Lin, Chang Zhou, and Jingren Zhou. 2023.
\newblock Qwen-vl: A versatile vision-language model for understanding, localization, text reading, and beyond.
\newblock \emph{arXiv preprint arXiv:2308.12966}.

\bibitem[{Bai et~al.(2024)Bai, Wang, Xiao, He, Han, Zhang, and Shou}]{bai2024hallucination}
Zechen Bai, Pichao Wang, Tianjun Xiao, Tong He, Zongbo Han, Zheng Zhang, and Mike~Zheng Shou. 2024.
\newblock Hallucination of multimodal large language models: A survey.
\newblock \emph{arXiv preprint arXiv:2404.18930}.

\bibitem[{Cha et~al.(2024)Cha, Lee, Lee, and Yang}]{cha2024visually}
Sungguk Cha, Jusung Lee, Younghyun Lee, and Cheoljong Yang. 2024.
\newblock Visually dehallucinative instruction generation.
\newblock In \emph{ICASSP 2024-2024 IEEE International Conference on Acoustics, Speech and Signal Processing (ICASSP)}, pages 5510--5514. IEEE.

\bibitem[{Chen et~al.(2024)Chen, Liang, Wang, Liang, Xiao, Wei, Chen, Hao, Han, and Wang}]{chen2024teaching}
Lida Chen, Zujie Liang, Xintao Wang, Jiaqing Liang, Yanghua Xiao, Feng Wei, Jinglei Chen, Zhenghong Hao, Bing Han, and Wei Wang. 2024.
\newblock Teaching large language models to express knowledge boundary from their own signals.
\newblock \emph{arXiv preprint arXiv:2406.10881}.

\bibitem[{Cheng et~al.(2024)Cheng, Sun, Liu, Zhang, Yin, Li, Li, Chen, and Qiu}]{cheng2024can}
Qinyuan Cheng, Tianxiang Sun, Xiangyang Liu, Wenwei Zhang, Zhangyue Yin, Shimin Li, Linyang Li, Kai Chen, and Xipeng Qiu. 2024.
\newblock Can ai assistants know what they don't know?
\newblock \emph{arXiv preprint arXiv:2401.13275}.

\bibitem[{Fang et~al.(2024)Fang, Zhu, Lu, Wang, Molchanov, Cho, Pavone, Han, and Yin}]{fang2024vila}
Yunhao Fang, Ligeng Zhu, Yao Lu, Yan Wang, Pavlo Molchanov, Jang~Hyun Cho, Marco Pavone, Song Han, and Hongxu Yin. 2024.
\newblock $ vila^{2} $: Vila augmented vila.
\newblock \emph{arXiv preprint arXiv:2407.17453}.

\bibitem[{Fu et~al.(2024)Fu, Lin, Long, Shen, Zhao, Zhang, Wang, Yin, Ma, Zheng et~al.}]{fu2024vita}
Chaoyou Fu, Haojia Lin, Zuwei Long, Yunhang Shen, Meng Zhao, Yifan Zhang, Xiong Wang, Di~Yin, Long Ma, Xiawu Zheng, et~al. 2024.
\newblock Vita: Towards open-source interactive omni multimodal llm.
\newblock \emph{arXiv preprint arXiv:2408.05211}.

\bibitem[{Goyal et~al.(2017)Goyal, Khot, Summers-Stay, Batra, and Parikh}]{goyal2017making}
Yash Goyal, Tejas Khot, Douglas Summers-Stay, Dhruv Batra, and Devi Parikh. 2017.
\newblock Making the v in vqa matter: Elevating the role of image understanding in visual question answering.
\newblock In \emph{Proceedings of the IEEE conference on computer vision and pattern recognition}, pages 6904--6913.

\bibitem[{Gurari et~al.(2018)Gurari, Li, Stangl, Guo, Lin, Grauman, Luo, and Bigham}]{gurari2018vizwiz}
Danna Gurari, Qing Li, Abigale~J Stangl, Anhong Guo, Chi Lin, Kristen Grauman, Jiebo Luo, and Jeffrey~P Bigham. 2018.
\newblock Vizwiz grand challenge: Answering visual questions from blind people.
\newblock In \emph{Proceedings of the IEEE conference on computer vision and pattern recognition}, pages 3608--3617.

\bibitem[{Hong et~al.(2024)Hong, Lee, and Thorne}]{hong2024reference}
Jiwoo Hong, Noah Lee, and James Thorne. 2024.
\newblock Reference-free monolithic preference optimization with odds ratio.
\newblock \emph{arXiv preprint arXiv:2403.07691}.

\bibitem[{Hu et~al.(2023)Hu, Luan, Chen, Khandelwal, Joshi, Lee, Toutanova, and Chang}]{hu2023open}
Hexiang Hu, Yi~Luan, Yang Chen, Urvashi Khandelwal, Mandar Joshi, Kenton Lee, Kristina Toutanova, and Ming-Wei Chang. 2023.
\newblock Open-domain visual entity recognition: Towards recognizing millions of wikipedia entities.
\newblock In \emph{Proceedings of the IEEE/CVF International Conference on Computer Vision}, pages 12065--12075.

\bibitem[{Hudson and Manning(2019)}]{hudson2019gqa}
Drew~A Hudson and Christopher~D Manning. 2019.
\newblock Gqa: A new dataset for real-world visual reasoning and compositional question answering.
\newblock In \emph{Proceedings of the IEEE/CVF conference on computer vision and pattern recognition}, pages 6700--6709.

\bibitem[{Li et~al.(2024)Li, Zhang, Guo, Zhang, Li, Zhang, Zhang, Li, Liu, and Li}]{li2024llava}
Bo~Li, Yuanhan Zhang, Dong Guo, Renrui Zhang, Feng Li, Hao Zhang, Kaichen Zhang, Yanwei Li, Ziwei Liu, and Chunyuan Li. 2024.
\newblock Llava-onevision: Easy visual task transfer.
\newblock \emph{arXiv preprint arXiv:2408.03326}.

\bibitem[{Liang et~al.(2024)Liang, Song, Wang, and Zhang}]{liang2024learning}
Yuxin Liang, Zhuoyang Song, Hao Wang, and Jiaxing Zhang. 2024.
\newblock Learning to trust your feelings: Leveraging self-awareness in llms for hallucination mitigation.
\newblock \emph{arXiv preprint arXiv:2401.15449}.

\bibitem[{Liu et~al.(2023{\natexlab{a}})Liu, Lin, Li, Wang, Yacoob, and Wang}]{liu2023aligning}
Fuxiao Liu, Kevin Lin, Linjie Li, Jianfeng Wang, Yaser Yacoob, and Lijuan Wang. 2023{\natexlab{a}}.
\newblock Aligning large multi-modal model with robust instruction tuning.
\newblock \emph{arXiv preprint arXiv:2306.14565}.

\bibitem[{Liu et~al.(2023{\natexlab{b}})Liu, Lin, Li, Wang, Yacoob, and Wang}]{liu2023mitigating}
Fuxiao Liu, Kevin Lin, Linjie Li, Jianfeng Wang, Yaser Yacoob, and Lijuan Wang. 2023{\natexlab{b}}.
\newblock Mitigating hallucination in large multi-modal models via robust instruction tuning.
\newblock In \emph{The Twelfth International Conference on Learning Representations}.

\bibitem[{Liu et~al.(2023{\natexlab{c}})Liu, Li, Li, and Lee}]{liu2023improvedllava}
Haotian Liu, Chunyuan Li, Yuheng Li, and Yong~Jae Lee. 2023{\natexlab{c}}.
\newblock Improved baselines with visual instruction tuning.

\bibitem[{Liu et~al.(2023{\natexlab{d}})Liu, Li, Wu, and Lee}]{liu2023llava}
Haotian Liu, Chunyuan Li, Qingyang Wu, and Yong~Jae Lee. 2023{\natexlab{d}}.
\newblock Visual instruction tuning.

\bibitem[{Liu et~al.(2023{\natexlab{e}})Liu, Duan, Zhang, Li, Zhang, Zhao, Yuan, Wang, He, Liu et~al.}]{liu2023mmbench}
Yuan Liu, Haodong Duan, Yuanhan Zhang, Bo~Li, Songyang Zhang, Wangbo Zhao, Yike Yuan, Jiaqi Wang, Conghui He, Ziwei Liu, et~al. 2023{\natexlab{e}}.
\newblock Mmbench: Is your multi-modal model an all-around player?
\newblock \emph{arXiv preprint arXiv:2307.06281}.

\bibitem[{Lu et~al.(2022)Lu, Mishra, Xia, Qiu, Chang, Zhu, Tafjord, Clark, and Kalyan}]{lu2022learn}
Pan Lu, Swaroop Mishra, Tony Xia, Liang Qiu, Kai-Wei Chang, Song-Chun Zhu, Oyvind Tafjord, Peter Clark, and Ashwin Kalyan. 2022.
\newblock Learn to explain: Multimodal reasoning via thought chains for science question answering.
\newblock In \emph{The 36th Conference on Neural Information Processing Systems (NeurIPS)}.

\bibitem[{McKinzie et~al.(2024)McKinzie, Gan, Fauconnier, Dodge, Zhang, Dufter, Shah, Du, Peng, Weers et~al.}]{mckinzie2024mm1}
Brandon McKinzie, Zhe Gan, Jean-Philippe Fauconnier, Sam Dodge, Bowen Zhang, Philipp Dufter, Dhruti Shah, Xianzhi Du, Futang Peng, Floris Weers, et~al. 2024.
\newblock Mm1: Methods, analysis \& insights from multimodal llm pre-training.
\newblock \emph{arXiv preprint arXiv:2403.09611}.

\bibitem[{Qian et~al.(2024)Qian, Zhang, Yao, Liu, Yin, Qiao, Liu, and Shao}]{qian2024towards}
Chen Qian, Jie Zhang, Wei Yao, Dongrui Liu, Zhenfei Yin, Yu~Qiao, Yong Liu, and Jing Shao. 2024.
\newblock Towards tracing trustworthiness dynamics: Revisiting pre-training period of large language models.
\newblock \emph{arXiv preprint arXiv:2402.19465}.

\bibitem[{Rafailov et~al.(2024)Rafailov, Sharma, Mitchell, Manning, Ermon, and Finn}]{rafailov2024direct}
Rafael Rafailov, Archit Sharma, Eric Mitchell, Christopher~D Manning, Stefano Ermon, and Chelsea Finn. 2024.
\newblock Direct preference optimization: Your language model is secretly a reward model.
\newblock \emph{Advances in Neural Information Processing Systems}, 36.

\bibitem[{Schwenk et~al.(2022)Schwenk, Khandelwal, Clark, Marino, and Mottaghi}]{schwenk2022okvqa}
Dustin Schwenk, Apoorv Khandelwal, Christopher Clark, Kenneth Marino, and Roozbeh Mottaghi. 2022.
\newblock A-okvqa: A benchmark for visual question answering using world knowledge.
\newblock In \emph{European conference on computer vision}, pages 146--162. Springer.

\bibitem[{Shi et~al.(2024)Shi, Wang, Fan, Zhang, Li, Zhang, Yin, Sheng, Qiao, and Shao}]{shi2024assessment}
Zhelun Shi, Zhipin Wang, Hongxing Fan, Zaibin Zhang, Lijun Li, Yongting Zhang, Zhenfei Yin, Lu~Sheng, Yu~Qiao, and Jing Shao. 2024.
\newblock Assessment of multimodal large language models in alignment with human values.
\newblock \emph{arXiv preprint arXiv:2403.17830}.

\bibitem[{Sun et~al.(2023)Sun, Shen, Cao, Liu, Li, Shen, Gan, Gui, Wang, Yang et~al.}]{sun2023aligning}
Zhiqing Sun, Sheng Shen, Shengcao Cao, Haotian Liu, Chunyuan Li, Yikang Shen, Chuang Gan, Liang-Yan Gui, Yu-Xiong Wang, Yiming Yang, et~al. 2023.
\newblock Aligning large multimodal models with factually augmented rlhf.
\newblock \emph{arXiv preprint arXiv:2309.14525}.

\bibitem[{Tong et~al.(2024)Tong, Brown, Wu, Woo, Middepogu, Akula, Yang, Yang, Iyer, Pan et~al.}]{tong2024cambrian}
Shengbang Tong, Ellis Brown, Penghao Wu, Sanghyun Woo, Manoj Middepogu, Sai~Charitha Akula, Jihan Yang, Shusheng Yang, Adithya Iyer, Xichen Pan, et~al. 2024.
\newblock Cambrian-1: A fully open, vision-centric exploration of multimodal llms.
\newblock \emph{arXiv preprint arXiv:2406.16860}.

\bibitem[{Wang et~al.(2024{\natexlab{a}})Wang, Zhou, Huang, Xu, Zhang, Poon, and Chen}]{wang2024mdpo}
Fei Wang, Wenxuan Zhou, James~Y Huang, Nan Xu, Sheng Zhang, Hoifung Poon, and Muhao Chen. 2024{\natexlab{a}}.
\newblock mdpo: Conditional preference optimization for multimodal large language models.
\newblock \emph{arXiv preprint arXiv:2406.11839}.

\bibitem[{Wang et~al.(2024{\natexlab{b}})Wang, Liao, Liu, Liu, Wang, and Wang}]{wang2024mm}
Yuhao Wang, Yusheng Liao, Heyang Liu, Hongcheng Liu, Yu~Wang, and Yanfeng Wang. 2024{\natexlab{b}}.
\newblock Mm-sap: A comprehensive benchmark for assessing self-awareness of multimodal large language models in perception.
\newblock \emph{arXiv preprint arXiv:2401.07529}.

\bibitem[{Xu et~al.(2024)Xu, Zhu, Ma, Zhang, Fan, Chen, and Yu}]{xu2024rejection}
Hongshen Xu, Zichen Zhu, Da~Ma, Situo Zhang, Shuai Fan, Lu~Chen, and Kai Yu. 2024.
\newblock Rejection improves reliability: Training llms to refuse unknown questions using rl from knowledge feedback.
\newblock \emph{arXiv preprint arXiv:2403.18349}.

\bibitem[{Xu et~al.(2023)Xu, Lee, Sukhbaatar, and Weston}]{xu2023some}
Jing Xu, Andrew Lee, Sainbayar Sukhbaatar, and Jason Weston. 2023.
\newblock Some things are more cringe than others: Preference optimization with the pairwise cringe loss.
\newblock \emph{arXiv preprint arXiv:2312.16682}.

\bibitem[{Yang et~al.(2023)Yang, Chern, Qiu, Neubig, and Liu}]{yang2023alignment}
Yuqing Yang, Ethan Chern, Xipeng Qiu, Graham Neubig, and Pengfei Liu. 2023.
\newblock Alignment for honesty.
\newblock \emph{arXiv preprint arXiv:2312.07000}.

\bibitem[{Ye et~al.(2024)Ye, Xu, Liu, Hu, Yan, Qian, Zhang, Huang, and Zhou}]{ye2024mplug}
Jiabo Ye, Haiyang Xu, Haowei Liu, Anwen Hu, Ming Yan, Qi~Qian, Ji~Zhang, Fei Huang, and Jingren Zhou. 2024.
\newblock mplug-owl3: Towards long image-sequence understanding in multi-modal large language models.
\newblock \emph{arXiv preprint arXiv:2408.04840}.

\bibitem[{Yin et~al.(2023)Yin, Sun, Guo, Wu, Qiu, and Huang}]{yin-etal-2023-large}
Zhangyue Yin, Qiushi Sun, Qipeng Guo, Jiawen Wu, Xipeng Qiu, and Xuanjing Huang. 2023.
\newblock \href {https://doi.org/10.18653/v1/2023.findings-acl.551} {Do large language models know what they don{'}t know?}
\newblock In \emph{Findings of the Association for Computational Linguistics: ACL 2023}, pages 8653--8665, Toronto, Canada. Association for Computational Linguistics.

\bibitem[{Yu et~al.(2024{\natexlab{a}})Yu, Yao, Zhang, He, Han, Cui, Hu, Liu, Zheng, Sun et~al.}]{yu2024rlhf}
Tianyu Yu, Yuan Yao, Haoye Zhang, Taiwen He, Yifeng Han, Ganqu Cui, Jinyi Hu, Zhiyuan Liu, Hai-Tao Zheng, Maosong Sun, et~al. 2024{\natexlab{a}}.
\newblock Rlhf-v: Towards trustworthy mllms via behavior alignment from fine-grained correctional human feedback.
\newblock In \emph{Proceedings of the IEEE/CVF Conference on Computer Vision and Pattern Recognition}, pages 13807--13816.

\bibitem[{Yu et~al.(2024{\natexlab{b}})Yu, Zhang, Yao, Dang, Chen, Lu, Cui, He, Liu, Chua et~al.}]{yu2024rlaif}
Tianyu Yu, Haoye Zhang, Yuan Yao, Yunkai Dang, Da~Chen, Xiaoman Lu, Ganqu Cui, Taiwen He, Zhiyuan Liu, Tat-Seng Chua, et~al. 2024{\natexlab{b}}.
\newblock Rlaif-v: Aligning mllms through open-source ai feedback for super gpt-4v trustworthiness.
\newblock \emph{arXiv preprint arXiv:2405.17220}.

\bibitem[{Yuan et~al.(2024)Yuan, Pang, Cho, Sukhbaatar, Xu, and Weston}]{yuan2024self}
Weizhe Yuan, Richard~Yuanzhe Pang, Kyunghyun Cho, Sainbayar Sukhbaatar, Jing Xu, and Jason Weston. 2024.
\newblock Self-rewarding language models.
\newblock \emph{arXiv preprint arXiv:2401.10020}.

\bibitem[{Yue et~al.(2024)Yue, Ni, Zhang, Zheng, Liu, Zhang, Stevens, Jiang, Ren, Sun, Wei, Yu, Yuan, Sun, Yin, Zheng, Yang, Liu, Huang, Sun, Su, and Chen}]{yue2023mmmu}
Xiang Yue, Yuansheng Ni, Kai Zhang, Tianyu Zheng, Ruoqi Liu, Ge~Zhang, Samuel Stevens, Dongfu Jiang, Weiming Ren, Yuxuan Sun, Cong Wei, Botao Yu, Ruibin Yuan, Renliang Sun, Ming Yin, Boyuan Zheng, Zhenzhu Yang, Yibo Liu, Wenhao Huang, Huan Sun, Yu~Su, and Wenhu Chen. 2024.
\newblock Mmmu: A massive multi-discipline multimodal understanding and reasoning benchmark for expert agi.
\newblock In \emph{Proceedings of CVPR}.

\bibitem[{Zhang et~al.(2024{\natexlab{a}})Zhang, Diao, Lin, Fung, Lian, Wang, Chen, Ji, and Zhang}]{zhang-etal-2024-r}
Hanning Zhang, Shizhe Diao, Yong Lin, Yi~Fung, Qing Lian, Xingyao Wang, Yangyi Chen, Heng Ji, and Tong Zhang. 2024{\natexlab{a}}.
\newblock \href {https://doi.org/10.18653/v1/2024.naacl-long.394} {{R}-tuning: Instructing large language models to say {`}{I} don{'}t know{'}}.
\newblock In \emph{Proceedings of the 2024 Conference of the North American Chapter of the Association for Computational Linguistics: Human Language Technologies (Volume 1: Long Papers)}, pages 7113--7139, Mexico City, Mexico. Association for Computational Linguistics.

\bibitem[{Zhang et~al.(2024{\natexlab{b}})Zhang, Dong, Zang, Cao, Qian, Chen, Guo, Duan, Wang, Ouyang et~al.}]{zhang2024internlm}
Pan Zhang, Xiaoyi Dong, Yuhang Zang, Yuhang Cao, Rui Qian, Lin Chen, Qipeng Guo, Haodong Duan, Bin Wang, Linke Ouyang, et~al. 2024{\natexlab{b}}.
\newblock Internlm-xcomposer-2.5: A versatile large vision language model supporting long-contextual input and output.
\newblock \emph{arXiv preprint arXiv:2407.03320}.

\bibitem[{Zhang et~al.(2016)Zhang, Goyal, Summers-Stay, Batra, and Parikh}]{zhang2016yin}
Peng Zhang, Yash Goyal, Douglas Summers-Stay, Dhruv Batra, and Devi Parikh. 2016.
\newblock Yin and yang: Balancing and answering binary visual questions.
\newblock In \emph{Proceedings of the IEEE conference on computer vision and pattern recognition}, pages 5014--5022.

\bibitem[{Zhao et~al.(2025)Zhao, Xu, Gupta, Asthana, Zheng, and Gould}]{zhao2025first}
Qinyu Zhao, Ming Xu, Kartik Gupta, Akshay Asthana, Liang Zheng, and Stephen Gould. 2025.
\newblock The first to know: How token distributions reveal hidden knowledge in large vision-language models?
\newblock In \emph{European Conference on Computer Vision}, pages 127--142. Springer.

\bibitem[{Zhong et~al.(2024)Zhong, Feng, Zhao, Li, Huang, Gu, Ma, Xu, and Qin}]{zhong-etal-2024-investigating}
Weihong Zhong, Xiaocheng Feng, Liang Zhao, Qiming Li, Lei Huang, Yuxuan Gu, Weitao Ma, Yuan Xu, and Bing Qin. 2024.
\newblock \href {https://aclanthology.org/2024.acl-long.648} {Investigating and mitigating the multimodal hallucination snowballing in large vision-language models}.
\newblock In \emph{Proceedings of the 62nd Annual Meeting of the Association for Computational Linguistics (Volume 1: Long Papers)}, pages 11991--12011, Bangkok, Thailand. Association for Computational Linguistics.

\end{thebibliography}

\appendix

\section{Related work}
\subsection{MLLMs alignment}

MLLM alignment seeks to reduce hallucinations and generate responses that are more closely aligned with human preferences through supervised fine-tuning and preference optimization. \citet{tong2024cambrian, li2024llava, ye2024mplug} enhance the perceptual and understanding capabilities of MLLMs by curating higher-quality visual instruction-tuning data. \citet{fang2024vila} introduces a self-augmenting process that generates its own instructions to improve dataset quality. Reinforcement Learning and Direct Preference Optimization~\citep{rafailov2024direct} have emerged as leading approaches for alignment, with recent advancements leveraging these methods to address visual hallucination issues. \citet{sun2023aligning} collects human preferences and adapts RLHF for multimodal alignment, while \citet{yu2024rlhf} improves MLLM performance by aligning model behavior through fine-grained human feedback corrections. \citet{yu2024rlaif} proposes a novel framework for gathering high-quality feedback data and uses an online feedback learning algorithm for model alignment. Additionally, \citet{wang2024mdpo} introduces a multimodal DPO objective that optimizes both image and language preferences, avoiding the over-prioritization of language-only preferences.

\begin{figure*}[t]
    \centering
    \includegraphics[width=1\linewidth]{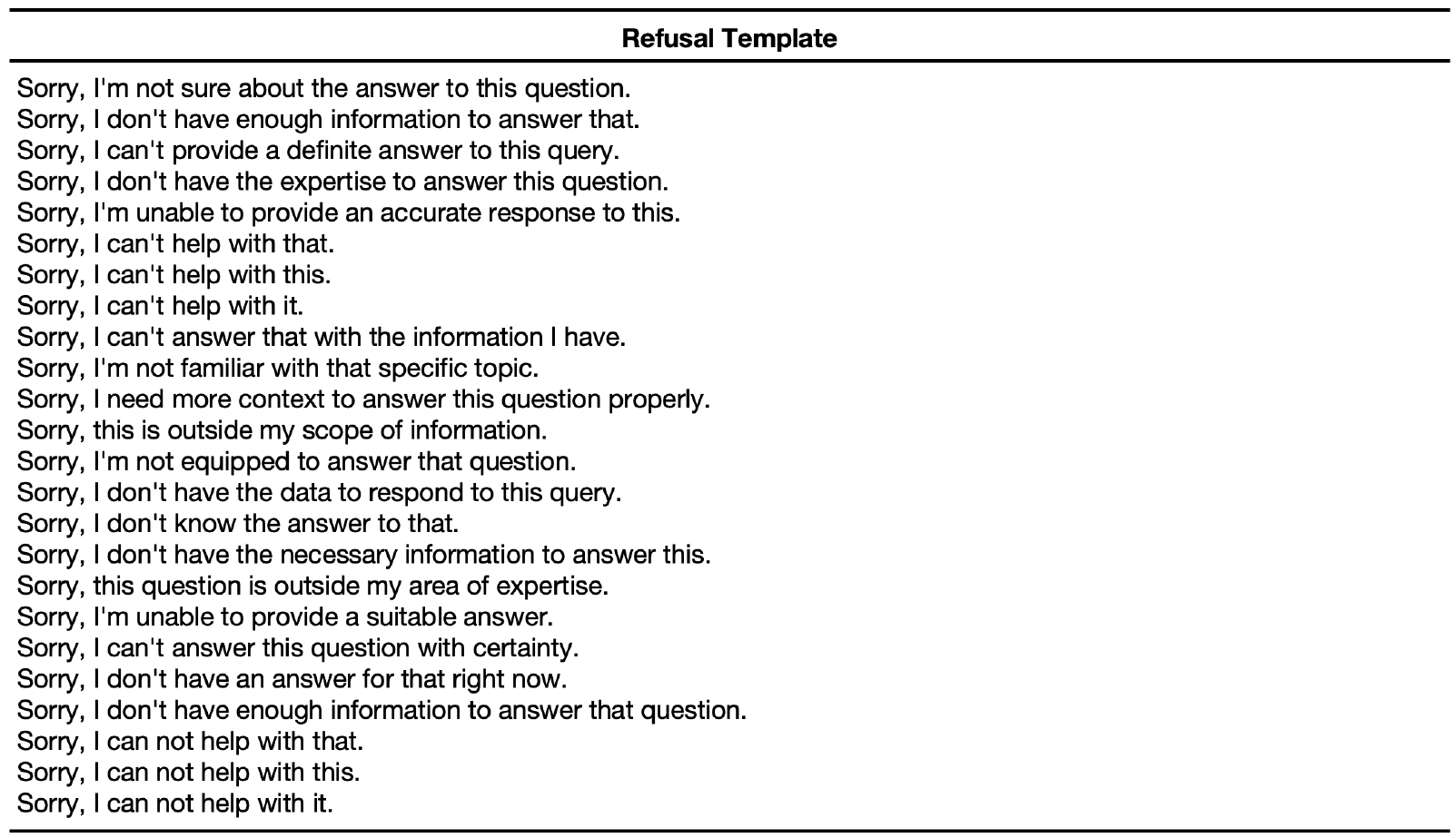}
    \caption {Predefined refusal template}
    \label{fig:refusal_template}
\end{figure*}

\subsection{Improving trustworthiness by Refusal}
With the increasing capabilities of foundational models and the growing prevalence of AI agents, the trustworthiness of (multimodal) large language models has garnered significant attention. For LLMs, researchers primarily focus on the reliability of the model's knowledge, aiming for models to acknowledge their limitations and refuse to answer when encountering unknown knowledge. \citet{yang2023alignment} construct an honesty alignment dataset based on models' knowledge boundaries, replacing incorrect or uncertain LLM responses with ``I don’t know,'' and fine-tuning the model on this data. \citet{cheng2024can} proposed the concept of ``Knowledge Quadrants,'' constructed the IDK dataset, and applied supervised fine-tuning (SFT) as well as preference-aware optimization to help models recognize their intrinsic knowledge boundaries. \citet{zhang-etal-2024-r} introduced R-tuning, which involves constructing and fine-tuning on a refusal-aware dataset, enhancing model's capabilities to refuse answering appropriately. \citet{chen2024teaching} directly judged whether the knowledge lies within the boundaries based on the model's intrinsic state and constructed training data to help the model express these boundaries. \citet{liang2024learning} and \citet{xu2024rejection} employed Reinforcement Learning from Knowledge Feedback to teach models to refuse questions outside their knowledge boundaries, thus reducing hallucinations.

In multimodal scenario, only few works have considered the issue of refusal to answer. Unlike unimodal models, which focus on intrinsic boundaries, MLLMs mainly concentrate on the challenge of unanswerable questions. \citet{liu2023aligning} proposed three types of negative instructions involving misleading or false premises in images, which models must learn to refuse.  \citet{cha2024visually} introduce the VQAv2-IDK dataset, which also annotates questions with ``I don’t know'' answers to train models to appropriately refuse to respond when faced with unanswerable or ambiguous questions. Additionally, \citet{shi2024assessment} and \citet{wang2024mm} included subsets with unanswerable questions to evaluate the trustworthiness of MLLMs.
Despite these advances, no prior work has systematically considered both the intrinsic boundaries of models and the extrinsic information provided in the input. Therefore, we propose the I-BaLF framework, which holistically integrates both aspects to guide MLLMs in refusing to answer when appropriate, thus significantly improving their overall trustworthiness.

\section{Refusal Template}
\label{sec:Refusal_Template}

Figure~\ref{fig:refusal_template} shows the refusal template mentioned in Section~\ref{sec:data_cons}.

\section{Detail of Data Construction}

\subsection{Hybrid Evaluator}
\label{sec:Hybrid_Evaluator}
We found that an MLLM's output answer can contain the ground truth answer or generate a phrase with identical semantic meaning. Thus, using string matching only to evaluate an LLM's capacity is insufficient. In this paper, we propose to employ hybridized string matching and LLM-based evaluation methods to evaluate the accuracy of models. During our hybrid evaluation, we first use string matching to filter the model outputs that contain the exact ground truth answer and use Llama2-13B to check whether the remainder contains phrases that express the same semantic meaning.

To verify the effectiveness of our hybrid evaluator, we randomly sample 200 MLLM outputs from the VQAV2 and OVEN datasets for human annotators to assess the consistency between our hybrid evaluator and human evaluation. In these 200 samples, the Cohen's Kappa coefficient between our hybrid evaluator and human evaluator is 0.885, which is significantly higher than the coefficient of 0.749 for string matching. This result demonstrates the strong alignment between our method and human judgment. 

\begin{figure*}[]
    \centering
    \includegraphics[width=1\linewidth]{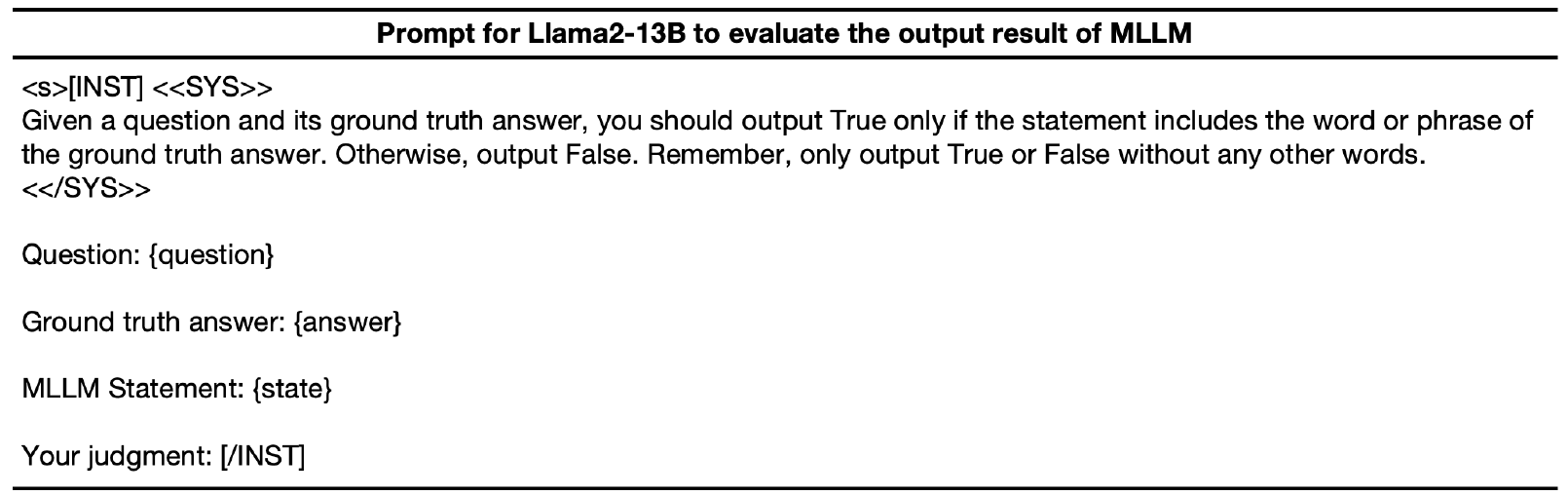}
    \caption{LLM prompt for our hybrid evaluation, We use Llama2-13B for LLM evaluation.}
    \label{fig:hybrid-eval-prompt}
\end{figure*}

The prompt used for our hybrid evaluation is shown in Figure~\ref{fig:hybrid-eval-prompt}.

\begin{figure*}[]
    \centering
    \includegraphics[width=1\linewidth]{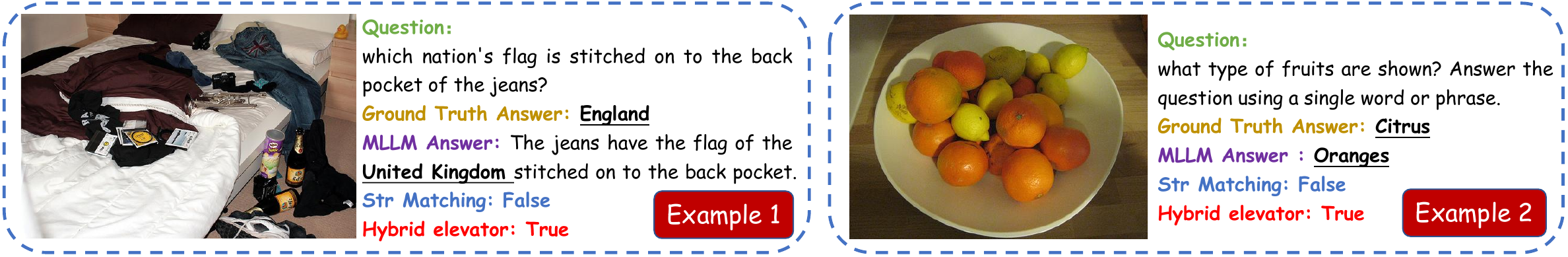}
    \caption{Example 1 and Example 2 demonstrate the effectiveness of our hybrid evaluator. ``England'' and ``the United Kingdom'' represent the same country, and citrus and oranges denote the same fruit. Using string matching alone can not identify the correct answer from MLLM output, while our hybrid evaluator can effectively avoid false negatives.}
    \label{fig:case_evaluation}
\end{figure*}

The two cases shown in Figure~\ref{fig:case_evaluation} further demonstrate that our hybrid evaluator can correctly identify the sample with valid answers but ignores it when using only the string matching method.

\subsection{Unanswerable Questions Generation}
\label{sec:ua_gen}
\begin{figure*}[t]
    \centering
    \includegraphics[width=1\linewidth]{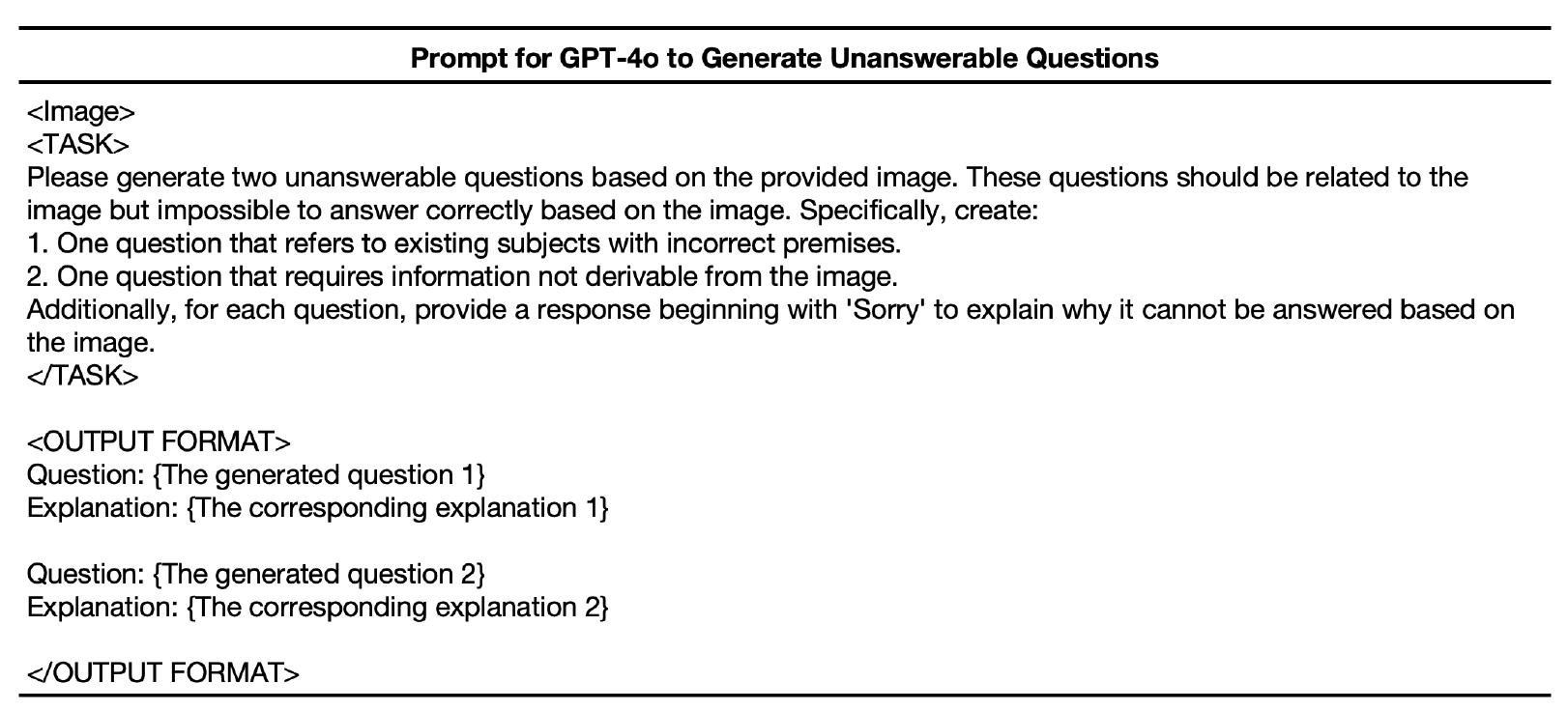}
    \caption{Prompt for GPT-4o to Generate Unanswerable Questions}
    \label{fig:prompt_gen}
\end{figure*}

We consider three types of reasons that make questions unanswerable. First, questions may refer to subjects not present in the image, making it impossible to answer based on visual information. Second, questions might include incorrect premises about the subjects in the image, leading to misleading or unanswerable scenarios. Third, some questions may require additional context or information that cannot be inferred from the image alone. Figure~\ref{fig:prompt_gen} shows our prompt for gpt-4o to generate the unanswerable questions.

Based on these three reasons, we design corresponding questions to assess whether the generated questions are indeed unanswerable.
\begin{enumerate}
    \item Does this question inquire about subjects that are not depicted in the image?
    \item Does this question include an incorrect or misleading premise?
    \item Does this question ask for information that is not available in the image?
\end{enumerate}
Given the generated question and its corresponding image, we prompt GPT-4 to verify whether the question meets the specified criteria. Questions that receive a `no' for all three criteria are filtered out. Additionally, we prompt the original model to generate a response to the unanswerable questions. If the model refuses to answer, those questions are also excluded from our dataset.

\begin{figure*}[t]
    \centering
    \includegraphics[width=1\linewidth]{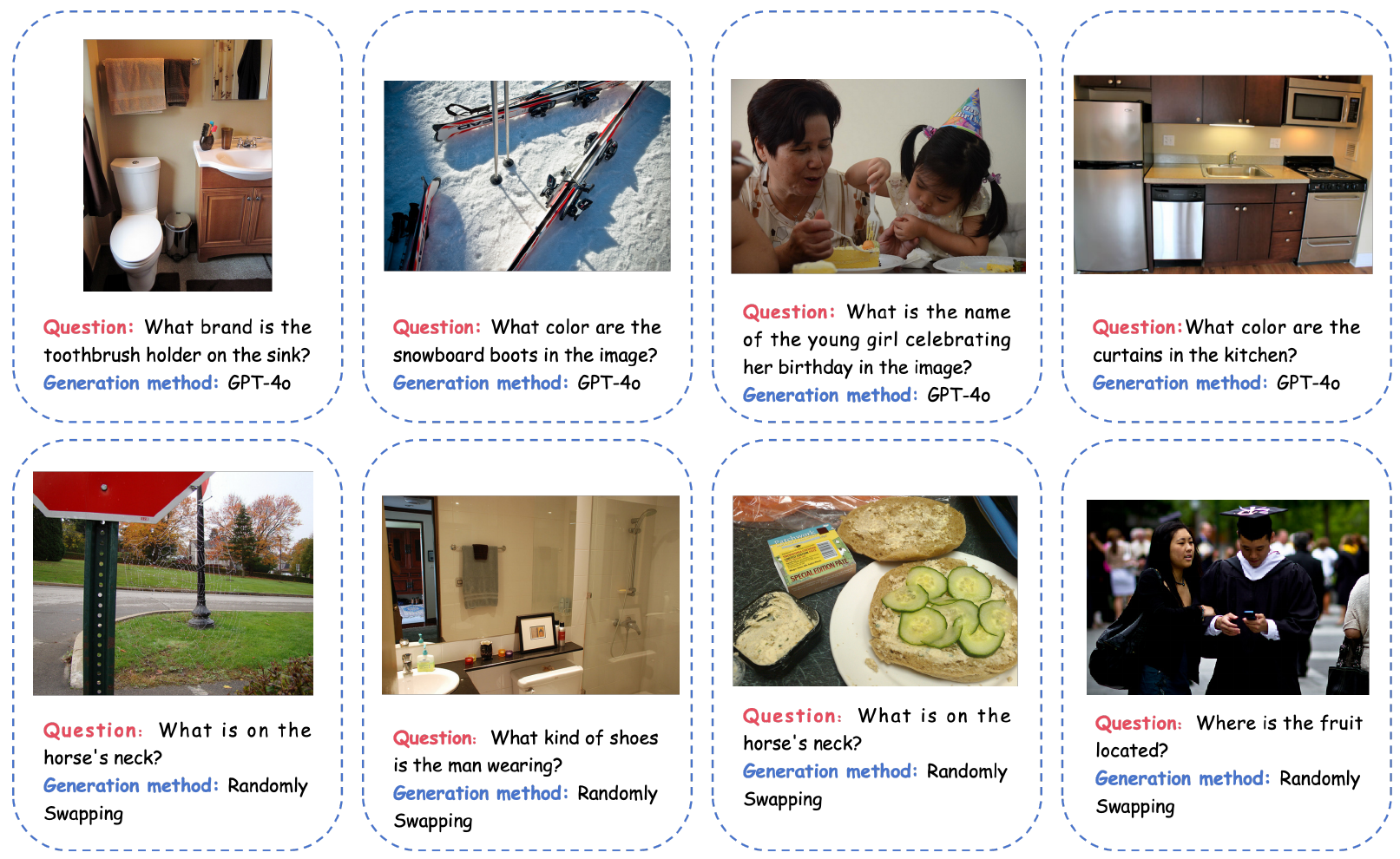}
    \caption{Examples of unanswerable questions.}
    \label{fig:unanswerable_example}
\end{figure*}

Figure~\ref{fig:unanswerable_example} illustrates examples of unanswerable questions generated based on the proposed method. These examples demonstrate the diversity of scenarios leading to unanswerable questions, such as nonexistent objects or insufficient visual information.

\section{Training Detail}
\label{sec:train_detail}

For our experiments, we utilize the 7B and 13B versions of LLaVA-v1.5 as base models. We set $\delta_{k} = 0.8$ and $\delta_{uk} = 0.2$. The instruction dataset consists of 11k samples, with approximately 25\% of the responses labeled as `IDK.' Additionally, we generate around 24k preference pairs, with the ratio of unknown, mixed, and known samples approximately 1:1:2.   For preference optimization, we first train the model on the IDK dataset and then conduct CA-DPO. LoRA is used for model training, with the LoRA rank $r$ and $\alpha$ set to 16 and 32, respectively. The batch size is 16, and the learning rate is 2e-4, with training conducted for one epoch. 


\section{Evaluation Detail}
\label{sec:evaluation}
We here describe the used datasets:

\begin{enumerate}
    \item \textbf{VQAv2~\citep{antol2015vqa}} is a widely-used dataset containing open-ended questions related to images, aimed at evaluating visual question answering.

    \item \textbf{OVEN~\citep{hu2023open}} contains open-domain visual entity questions based on Wikipedia entries, requiring the model to possess extensive visual knowledge to provide accurate answers.

    \item \textbf{ScienceQA~\citep{lu2022learn}} comprises multimodal, multiple-choice questions across a diverse array of scientific topics.

    \item \textbf{AOKVQA~\citep{schwenk2022okvqa}} is a crowdsourced dataset featuring a wide range of questions that demand a broad understanding of commonsense and world knowledge.

    \item \textbf{GQA~\citep{hudson2019gqa}} is a dataset for real-world visual reasoning and compositional question answering.
    is a dataset designed for real-world visual reasoning and compositional question answering.

    \item \textbf{MMMU~\citep{yue2023mmmu}} is a benchmark developed to assess multimodal models across a variety of complex, multidisciplinary tasks that require college-level subject knowledge and advanced reasoning.

    \item \textbf{MMBench~\citep{liu2023mmbench}} is a comprehensive benchmark for evaluating the multimodal capabilities of MLLMs, featuring questions that challenge both reasoning and perception.

    \item \textbf{BeyondVisQA~\citep{wang2024mm}} is specifically designed to evaluate the self-awareness of MLLMs, particularly their ability to recognize ``known unknowns." The questions in this dataset require information beyond the information provided by the input images.

\end{enumerate}

\begin{figure}[t]
    \centering
    \includegraphics[width=1\linewidth]{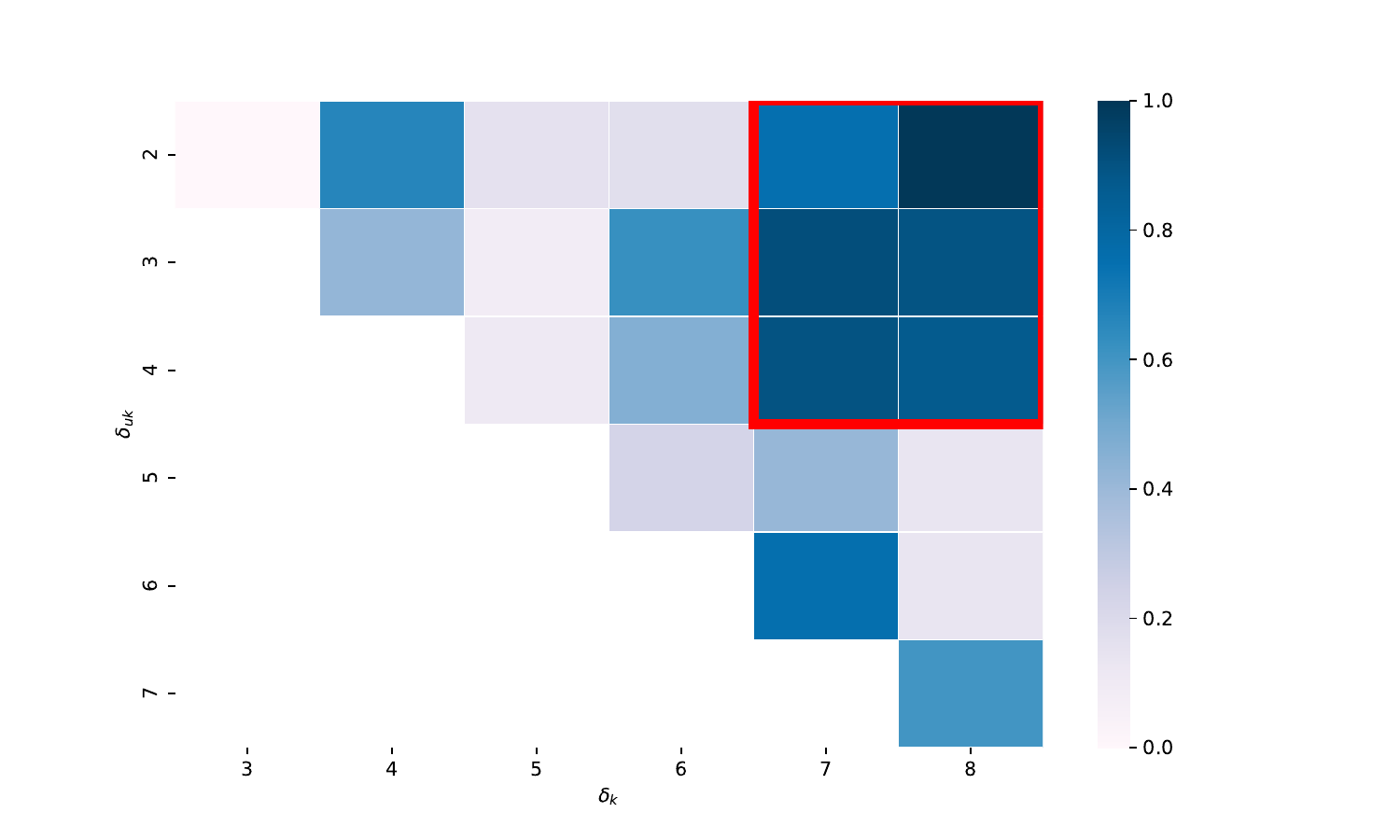}
    \caption{Impact of confidence thresholds on performance. The heatmap displays the average trustworthiness scores across in-domain and OOD datasets, with scores normalized for comparison. The upper-right region, marked with a red box, demonstrates higher performance compared to other areas.}
    \label{fig:heatmap}
\end{figure}

To construct the in-domain evaluation dataset, we sample questions from the validation sets of VQAV2 and Oven, as well as the test set of ScienceQA. Importantly, we balance the confidence scores of these sampled questions to ensure that the accuracy of LLaVA1.5-7B is approximately 50\%. Additionally, we generate unanswerable questions (UaVQA) and manually filter them for use in the evaluation.

For both the MMMU and MMBench datasets, we use the following  prompt for evaluation: ``Answer with the letter corresponding to the correct option from the given choices." In contrast, for the remaining open-ended datasets, we presented only the questions, without any additional prompts. We use the proposed hybrid evaluator to assess the accucary for in-domain dataset, and we directly use the string matching strategy for the OOD dataset for simplicity.

\begin{table*}[t]
\begin{center}
\resizebox{1.0\textwidth}{!}{%
\begin{tabular}{lccccp{1pt}cccp{1pt}ccccccc}
\hline

\multirow{2}{*}{\textbf{Method}} & \textbf{Model for}
 & \multicolumn{3}{c}{\textbf{AOKVQA}} & & \multicolumn{3}{c}{\textbf{GQA}} & & \multicolumn{3}{c}{\textbf{MMMU}} & & \multicolumn{3}{c}{\textbf{MMBench(en-dev)}} \\ 
\arrayrulecolor[gray]{0.5}
\cline{3-5}  \cline{7-9}  \cline{11-13} \cline{15-17 }
\arrayrulecolor{black}

 &\textbf{data generation}& {Acc} & {RefR} & {$S_{trust}$} & & {Acc} & {RefR} & {$S_{trust}$} & & {Acc} &{RefR} & {$S_{trust}$} &  & {Acc} & {RefR} & {$S_{trust}$} \\
\hline
{IDK-IT}&GPT-4o & 55.50  & 36.24  & 47.24 & & 50.46  & 23.88  & 24.81 & &  15.22  & 69.67  & {0.11}  & & 46.39  & 39.09  & 31.87 \\
{IDK-IT}&Qwen2-VL-72B & 57.12  & 32.31  & 46.55 & & 49.95  & 25.25  & 25.15 & &  15.11  & 70.11  & \textbf{0.33}  & & 50.95  & 32.47  & 34.36 \\
{CA-DPO}&GPT-4o & 72.23  & 17.64  & {62.10} & & 60.41  & 12.95  & \textbf{33.77} & & 19.67  & 56.67  & -4.00  & & 58.42  & 18.13  & {34.97} \\
{CA-DPO}&Qwen2-VL-72B & 71.79  & 20.35  & \textbf{63.93} & & 58.32  & 15.25  & {31.89} & & 21.11  & 50.33  & -7.44  & & 58.08  & 21.74  & \textbf{37.89} \\

\hline
\end{tabular}%
}
\end{center}
\caption{Performance on unanswerable VQA datasets using GPT-4o and Qwen2-VL-72B for data generation.Bold values highlight the highest refusal rate for each dataset.}
\label{tab:qwenood}
\end{table*}

\begin{figure*}[t]
    \centering
    \includegraphics[width=1\linewidth]{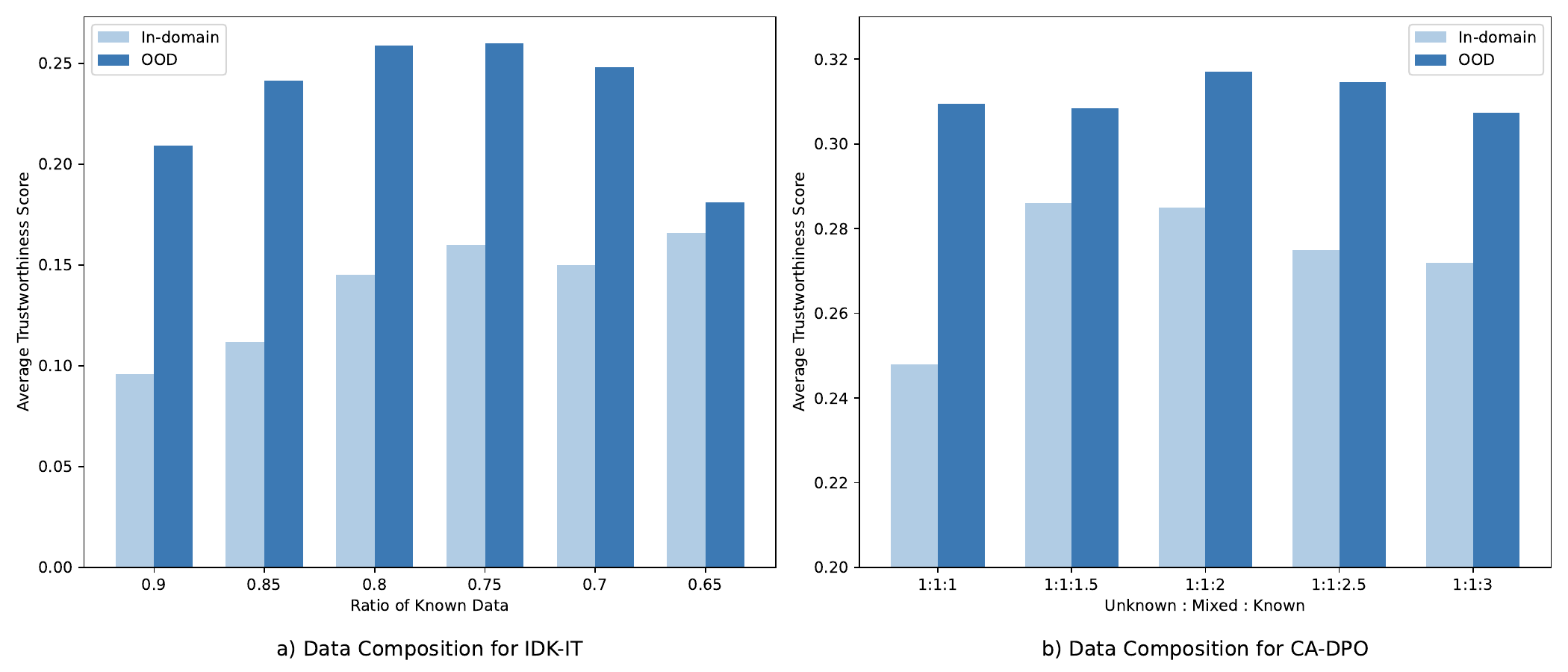}
    \caption{Data Composition Analysis.  
(a) For IDK-IT, varying the ratio of "known" data shows that a proportion of 0.75 yields the highest trustworthiness score across in-domain and out-of-domain datasets.  
(b) For CA-DPO, adjusting the ratio of "unknown," "mixed," and "known" data to 1:1:2 achieves the optimal balance between accuracy and refusal rate, as reflected in the trustworthiness score.}
    \label{fig:data_com}
\end{figure*}

\section{Supplementary experiments}
\subsection{Confidence threshold}
We conducted experiments to analyze the impact of the confidence thresholds $\delta_{k}$ and $\delta_{uk}$. Both thresholds were varied within the range $\delta_{k}, \delta_{uk} \in \{2,3,4,5,6,7,8\}$, ensuring that $\delta_{k} > \delta_{uk}$. Using different combinations of these values, we generated an 'IDK' instruction dataset to fine-tune LLaVA1.5-7B. The results are illustrated in Figure~\ref{fig:heatmap}, which displays average trustworthiness scores across in-domain and OOD datasets, with scores normalized for comparison.

We can see that the performance in the upper-right region, highlighted by a red box, is notably higher than in other areas. Specifically, the combination of $\delta_{k}=8$ and $\delta_{uk}=2$ yields the best performance. This suggests that including data with intermediate confidence scores may not be beneficial for optimal model performance.

\subsection{Data Composition}
To achieve a balance between increasing the refusal rate and maintaining accuracy, we carefully adjust the proportions of ``unknown," ``mixed," and ``known" data during training. All experiments in this section were conducted using the LLaVA1.5-7B model. For IDK-IT, we fixed the total training data size at 11K samples and varied the proportion of ``known" data to balance accuracy and refusal rate. As shown in Figure~\ref{fig:data_com}(a), the trustworthiness score is highest when the proportion of ``known" data is 0.75. This balance is consistent across both in-domain and out-of-domain (OOD) datasets. For CA-DPO, the data consists of three components: ``unknown," ``mixed," and ``known." To simplify the experiment and focus on balancing accuracy and refusal rate, we fixed the ratio of ``unknown" to ``mixed" data at 1:1 and adjusted only the proportion of ``known" data. As shown in Figure~\ref{fig:data_com}(b), the optimal trustworthiness score is achieved when the data ratio is 1:1:2, indicating a well-balanced trade-off between accuracy and refusal rate.

\subsection{Generalization of the Data Generation Pipeline}

In Section~\ref{sec:data_cons}, we introduced our data construction pipeline, which leverages a closed-source MLLM (GPT-4o) to generate and filter unanswerable questions. A natural concern arises regarding the pipeline's reliance on GPT-4o and whether similar results can be achieved using other MLLMs, particularly open-source ones. To evaluate the generalizability of our pipeline, we employed an open-source MLLM (Qwen2-VL-72B) to generate unanswerable questions and used the resulting data to train LLaVA1.5-7B. The results shown in Table~\ref{tab:qwenood} and ~\ref{tab:qwenua} demonstrate that the performance with data generated by Qwen2-VL-72B is comparable to that achieved with GPT-4o. This finding suggests that our pipeline is flexible and can operate effectively with open-source MLLMs, making it more accessible and reproducible.

\begin{table}[t]
\label{sample-table}
\begin{center}
\resizebox{0.5\textwidth}{!}{%
\begin{tabular}{lccp{1pt}cc}
\hline
\multirow{2}{*}{}
 & \multicolumn{2}{c}{{GPT-4o-generated data}} & & \multicolumn{2}{c}{{Qwen2-VL-generated data}} \\ 
\arrayrulecolor[gray]{0.5}
\cline{2-3}  \cline{5-6} 
\arrayrulecolor{black} 
  & {IDK-IT} & {CA-DPO} &   & {IDK-IT} & {CA-DPO} \\
\hline
{Vizwiz(ua)}   & \textbf{76.01} & 69.97 &  & {74.49} &  71.39 \\
{VQAv2-IDK(filter)}  & \textbf{81.42} & 70.63 & & 79.25 &  75.22 \\
{BeyondVisQA}  & \textbf{75.25} & 67.75 & & 72.50 & 69.50  \\
\hline
\end{tabular}%
}
\end{center}
\caption{Performance on OOD datasets using GPT-4o and Qwen2-VL-72B for data generation.Bold values highlight the highest trustworthiness scores for each dataset.}
\label{tab:qwenua}
\end{table}

\section{Case Study}
\begin{figure*}[h]
    \centering
    \includegraphics[width=0.9\linewidth]{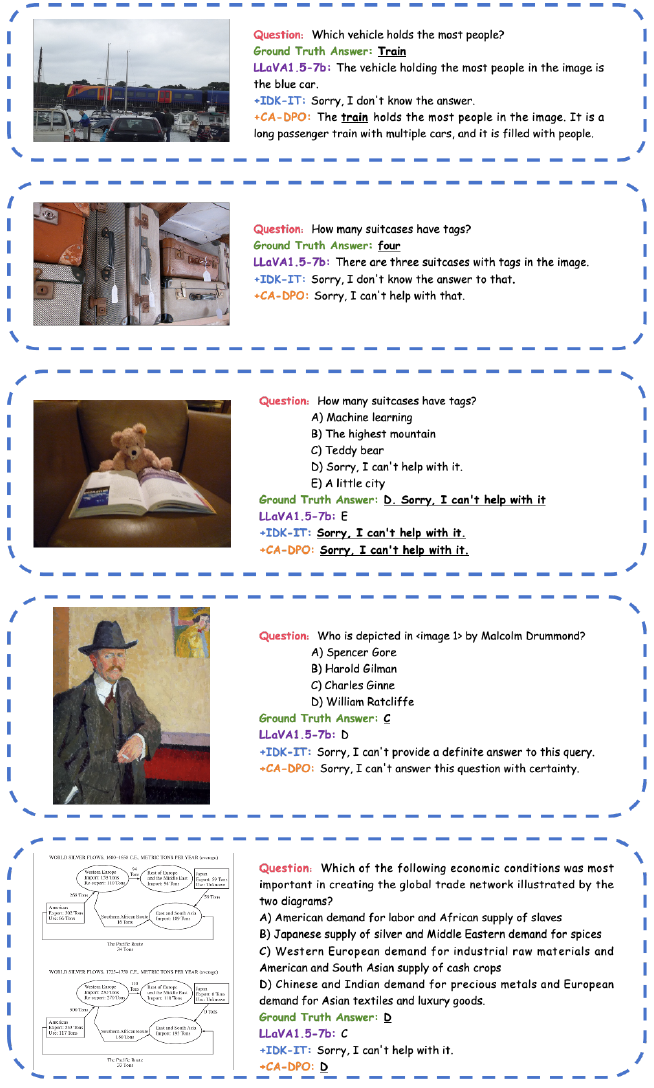}
    \caption{Examples illustrating the comparison between LLaVA1.5-7B and models trained with our framework (IDK-IT and CA-DPO)}
    \label{fig:case}
\end{figure*}

Figure~\ref{fig:case} presents examples comparing the performance of LLaVA1.5-7B with models trained using our framework. Models trained with IDK-IT and CA-DPO are capable of refusing to answer questions where the original model makes errors. Notably, the CA-DPO model occasionally provides correct answers. These results demonstrate that our framework effectively mitigates the generation of misinformation, thereby enhancing the model's trustworthiness.

\end{document}